\pdfoutput=1

\documentclass[11pt]{article}

\usepackage[final]{acl}

\usepackage{times}
\usepackage{latexsym}
\usepackage{enumitem}

\usepackage[T1]{fontenc}

\usepackage[utf8]{inputenc}

\usepackage{microtype}

\usepackage{inconsolata}

\usepackage{graphicx}
\usepackage{subfigure}
\usepackage{amssymb,amsthm,dsfont,mathrsfs}
\usepackage{amsmath}
\usepackage{setspace}
\usepackage{multirow}
\usepackage{helvet}
\usepackage{courier}
\usepackage{bm}
\usepackage{mdwlist}
\usepackage{booktabs}
\usepackage{algorithm}
\usepackage{algorithmic}
\usepackage{url}

\newcommand\blfootnote[1]{%
  \begingroup
  \renewcommand\thefootnote{}\footnote{#1}%
  \addtocounter{footnote}{-1}%
  \endgroup
}

\title{Unveiling and Consulting Core Experts in Retrieval-Augmented MoE-based LLMs}

\author{ 
 Xin Zhou\textsuperscript{\rm 1*$\ddagger$}, 
 Ping Nie\textsuperscript{\rm 2*}, 
 Yiwen Guo\textsuperscript{\rm 6$\dagger$},
 Haojie Wei\textsuperscript{\rm 2},
 Zhanqiu Zhang\textsuperscript{\rm 2} \\
 \textbf{Pasquale Minervini\textsuperscript{\rm 5},
Ruotian Ma\textsuperscript{\rm 1}, 
 Tao Gui\textsuperscript{\rm 3,4$\dagger$}, 
  Qi Zhang\textsuperscript{\rm 1,4$\dagger$},
  Xuanjing Huang\textsuperscript{\rm 1,4}}  \\
{$^1$School of Computer Science, Fudan University, Shanghai, China} {$^2$ LightSpeed Studios, Tencent} \\
  {$^3$ Institute of Modern Languages and Linguistics, Fudan University, Shanghai, China} \\
  {$^4$ Key Laboratory of Intelligent Information Processing, Fudan University, Shanghai, China} \\
  {$^5$ School of Informatics and ELLIS, University of Edinburgh} 
  {$^6$ Independent Researcher}
  \\
  \texttt{{ \{xzhou20, qz\}@fudan.edu.cn}}
  }

\begin{document}
\maketitle
\begin{abstract}
Retrieval-Augmented Generation (RAG) significantly improved the ability of Large Language Models (LLMs) to solve knowledge-intensive tasks.
While existing research seeks to enhance RAG performance by retrieving higher-quality documents or designing RAG-specific LLMs, the internal mechanisms within LLMs that contribute to the effectiveness of RAG systems remain underexplored.
In this paper, we aim to investigate these internal mechanisms within the popular Mixture-of-Expert (MoE)-based LLMs and demonstrate how to improve RAG by examining expert activations in these LLMs.
Our controlled experiments reveal that several core groups of experts are primarily responsible for RAG-related behaviors.
The activation of these core experts can signify the model's inclination towards external/internal knowledge and adjust its behavior.
For instance, we identify core experts that can (1) indicate the sufficiency of the model's internal knowledge, (2) assess the quality of retrieved documents, and (3) enhance the model's ability to utilize context.
Based on these findings, we propose several strategies to enhance RAG's efficiency and effectiveness through expert activation.
Experimental results across various datasets and MoE-based LLMs show the effectiveness of our method.
\end{abstract}
\blfootnote{$^*$Equal contribution.}
\blfootnote{$^\dagger$Corresponding authors.}
\blfootnote{$^\ddagger$Work done during the internship at Tencent LightSpeed Studios.}

\section{Introduction}
Retrieval-Augmented Generation~\cite[RAG,][]{lewis2020retrieval,gao2024retrievalaugmented,ding2024survey} has shown significant achievements in enhancing Large Language Models~\cite[LLMs,][]{brown2020language,chowdhery2023palm,touvron2023llama}.
By retrieving relevant documents from external knowledge bases and incorporating them into the context, RAG allows LLMs to access query-relevant and up-to-date information, thereby improving their performance on a variety of knowledge-intensive NLP tasks \cite{lozano2023clinfoai,kang2023deficiency}.
\begin{figure}[t]
\centering
  \includegraphics[width=1.0\linewidth]{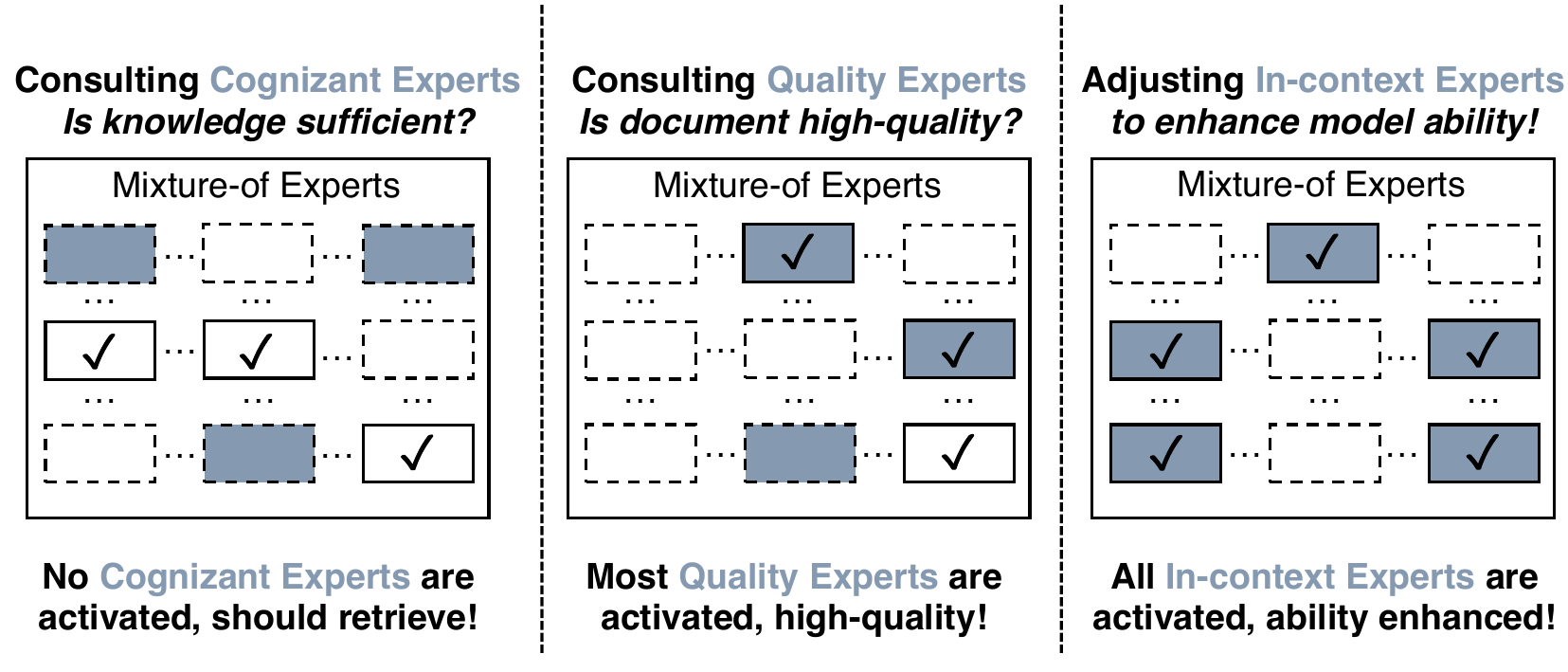}\vspace{-0.4cm}
  \caption{
  Three types of core experts identified in our experiments and their applications in RAG scenarios.
  Blue colors represent core experts, while solid lines and $\checkmark$ indicate activated experts.
  Cognizant experts indicate whether knowledge is sufficient; Quality experts evaluate the quality of retrieval documents; In-context experts enhance the LLM's ability.
}
\vspace{-0.4cm}
 \label{fig:three_experts}
\end{figure}

Despite these achievements, RAG faces many challenges \cite{chen2024benchmarking}. 
 For instance, long retrieved documents introduce additional inference cost \cite{xu2023retrieval}, irrelevant or erroneous retrieved documents may lead to increased hallucinations \cite{shi2023large,mallen-etal-2023-trust}, and LLMs might not effectively utilize related information from context \cite{xie2023adaptive}. 
Although significant efforts have been made to improve the quality of retrieved documents \cite{xie2023adaptive,wang2023learning} and train specialized models for RAG \cite{asai2024selfrag,lin2023ra}, there is limited research examining RAG from the perspective of LLMs' internal mechanisms.

In this paper, we aim at paying more attention to the internal states of retrieval-augmented LLMs, focusing on Mixture-of-Expert (MoE)-based LLMs \cite{du2022glam,jiang2024mixtral} 
whose inner expert activations naturally reveal their internal states.
Specifically, MoE-based LLMs comprise a set of experts that are often activated differently depending on the input context. We believe that certain core experts within the models play a vital role in managing specific types of contexts and regulating model behaviors. As a consequence, examining the function of these experts in conjunction with RAG can enhance our understanding of RAG's benefits and provide insights into how it can be further improved to address the aforementioned challenges.

We present Contrastive Expert Activation Inspection (CEAI), a simple but effective method for inspecting internal mechanisms of MoE-based LLMs. CEAI works by comparing the activation of experts given contrastive contexts, which are designed to induce opposite model behaviors.
 
As illustrated in Figure \ref{fig:three_experts}, CEAI helps discover three types of RAG-related experts that exhibit unique activation patterns, namely cognizant experts, quality experts, and in-context experts. 
Identified by inspecting activations when the MoE-based LLMs generate correct versus incorrect answers, cognizant experts determine the sufficiency of the model's internal knowledge to response to user queries on its own. 
Quality experts and in-context experts, on the other hand, assess the quality of retrieved documents and adjust the model's information utilization abilities from the context, respectively.

Based on these findings, we propose a training-free adaptive RAG method by investigating and manipulating the activation of experts in MoE-based LLMs. 
Specifically, we let the activation of cognizant experts and quality experts serve as indicators of unnecessary retrieval and low-quality documents. By avoiding unnecessary retrieval and filtering out low-quality documents, the efficiency of RAG can be improved. 
We additionally enhance the model's ability to utilize contextual information by adjusting the activation of in-context experts.
Moreover, we design data recipes and metrics to enhance the comprehensiveness of adaptive RAG evaluation.
Experimental results across various datasets show the advantages of our method. Our codes are publicly available at \url{https://github.com/xzhou20/Expert-RAG}.

Our contribution can be summarized as follows:
\begin{itemize}
    \item 
    We propose CEAI, a method for detecting core experts that manage specific context types and model behaviors in MoE-based LLM. 
    \item We explore the impact of specific experts to the RAG process, discovering three types of RAG-related experts. 
    These experts help determine knowledge sufficiency, assess the quality of retrieved documents, and enhance the model's ability to utilize context, showing potential for improving RAG.
    \item Based on our findings, we take explicit advantage of these core experts to enhance the effectiveness and adaptivity of RAG with MoE-based LLMs. 
    We verify the effectiveness of our method with comprehensive evaluation.
\end{itemize}

\begin{figure*}[t]
    \centering
    \includegraphics[width=0.9\linewidth]{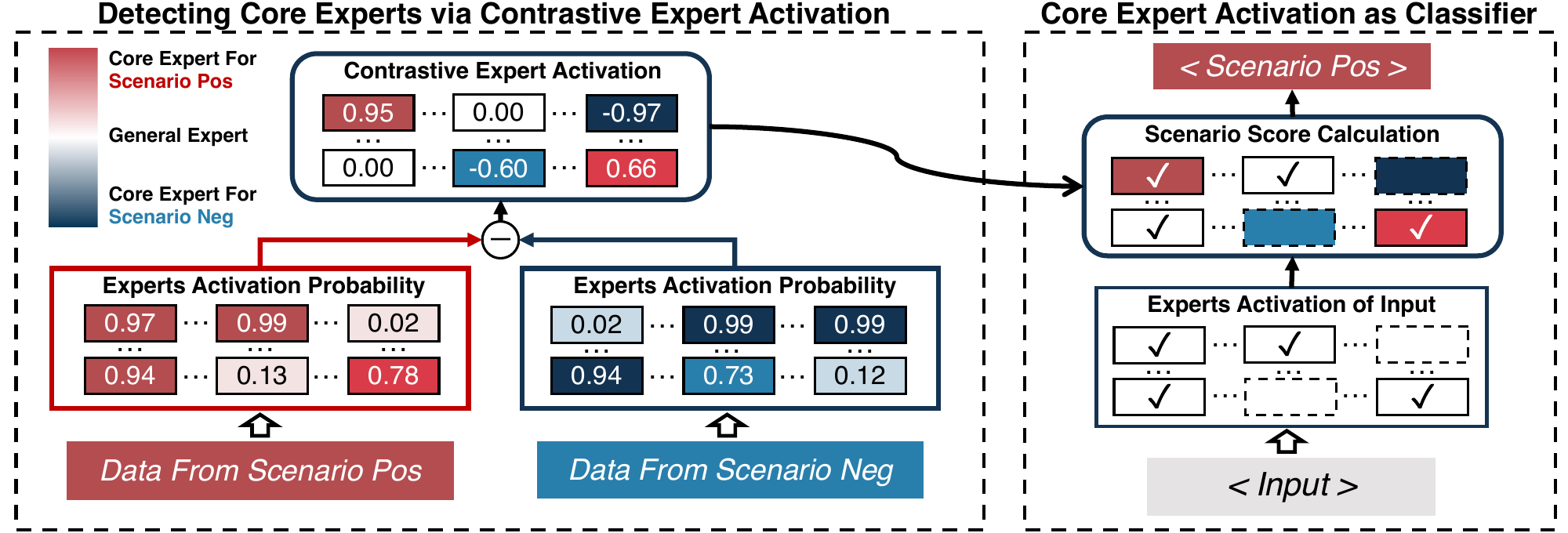}
    \caption{
    An overview of our methods. To detect core experts, we use data from Pos and Neg scenarios to inspect MoE-based LLMs, collecting experts that are frequently activated only in corresponding scenarios. 
    By comparing expert activation in contrastive scenarios, we identified core experts that are highly activated in specific scenarios. 
    The activation of these core experts can be used as classifiers to predict the scenario for new input.}
\vspace{-0.4cm}
    \label{fig:method}
\end{figure*}

\section{Method}

\subsection{Expert Activation in MoE}
\label{sec:moe}
The MoE architecture replaces the Feed-Forward Network (FFN) module with the MoE module in each transformer layer.
The MoE module typically consists of a routing network and multiple FFNs, each FFN module acting as an expert.
During LLM's forward phase, the routing network selectively feeds each token to the most appropriate experts, and only the selected experts are activated and contribute to the computation. 
The output of the MoE module is the weighted sum of the activated experts. 
Formally, given an MoE module with $N$ experts and an input token representation $\mathbf{h}$, the output of the MoE module in $i$-th layer is:
\begin{equation}
    \text{MoE}_i(\mathbf{h}) = \sum_{j=1}^{N} g_{i,j}(\mathbf{h})  e_{i,j}(\mathbf{h}),
\end{equation}
where $e_{i,j}(\mathbf h)$ is the output of the $j$-th expert in $i$-th layer, and $g_{i,j}(\mathbf h)$ is the gating value computed by the routing network. 
Typically, each token activates the top-$k$ experts per layer, making the remaining $g_{i,j}(\mathbf{h})$ zero. This indicates that those experts are not suitable for the current context.
\subsection{Contrastive Expert Activation Inspection}
Since experts within the MoE are dynamically activated according to the demand of the context, 
we hypothesize that there are some core experts primarily responsible for specific contexts and model behaviors. 
To identify these core experts, we propose a straightforward yet effective method, Contrastive Expert Activation Inspection (CEAI), which detects core experts by comparing the expert activation frequency across contrastive scenarios.

An overview of CEAI is shown in Figure~\ref{fig:method}. 
We define a scenario as a set of input prompts that induce consistent model behavior. 
Contrastive scenarios include a positive scenario and a negative scenario, representing two types of input prompts designed to elicit opposite model behaviors.  
For example, a positive scenario can be that input prompt includes external documents, while a negative scenario excludes external documents. 
MoE LLMs require different abilities to handle these two contrastive scenarios, thus activating different experts to exhibit opposite behaviors. 
By comparing the activation frequencies of experts, we can exclude general experts that are activated in both scenarios, thereby highlighting the core experts more likely to be activated in their respective scenarios.

Given two dataset $D_\text{pos}$ and $D_\text{neg}$ 
representing data from contrastive scenarios, let $\mathbf{h}_i=f(X)$ denote the input representation to the $i$-th layer MoE module for input prompt $X$.
We introduce the concept of activation probability of the \(j\)-th expert on the \(i\)-th layer for scenario pos as:
\begin{equation}
\label{eq:p_d}
    P^{\text{pos}}_{e_{i,j}} = \frac{1}{|D_\text{pos}|} \sum_{X \in D_\text{pos}}{\mathbb{I}(g_{i,j}(\mathbf{h}_i))},
\end{equation}
where $\mathbb{I}(g_{i,j}(\mathbf{h}_i)) \rightarrow \{0,1\}$ indicates whether expert $e_{i,j}$ is activated for the $\mathbf{h}_i$. 
We take the last position of input prompt to calculate $\mathbf{h}_i$ for all experiments and show explanation in Appendix \ref{appendix:CEAI}.

We then introduce the contrastive activation probabilities, which is the difference between activation probabilities in two contrastive scenarios:
\begin{equation}
\label{eq:final_expert}
    \Delta P_{i,j} = P^{\text{pos}}_{e_{i,j}} - P^{\text{neg}}_{e_{i,j}},
\end{equation}
where $\Delta P_{i,j} > 0$ indicates a higher activation probability for expert $e_{i,j}$ in the pos scenario compared to the neg scenario, suggesting that this expert is more responsible for the pos scenario. A negative $\Delta P_{i,j}$ suggests the opposite. 

\paragraph{Expert Activation Pattern for Classification}
\label{sec:activation_as_classifier}
We can use expert activation as a classifier to predict the type of scenarios, thereby determining the act of model for improving RAG. 
For example, if the expert activation can predict the internal knowledge of the model is sufficient for the current query, we can 
avoid unnecessary retrieval and enhance the efficiency of RAG.
For this aim, we introduce a \textbf{Scenario Score} for classifying the scenarios based on the activation of core experts. 
Given contrastive activation probabilities  obtained from Equation \ref{eq:final_expert}, the scenario score of any input is calculated as:
\begin{equation}
\label{eq:score}
    \text{Scenario Score}  = \sum_{i=1}^{L} \sum_{j=1}^{N} \Delta P_{i,j} \cdot \mathbb{I}(g_{i,j}(\mathbf{h}_i)),
\end{equation}
where \(L\) is the number of layers, \(N\) is the number of experts per layer, 
and $ \mathbb{I}(g_{i,j}(\mathbf{h}_i))$ is the indicator function same to Equation \ref{eq:p_d}. 
A positive scenario score indicates a higher inclination towards the positive scenario, while a negative score indicates a tendency towards the negative scenario.
This method has several variations. For instance, we can limit the calculation to the top and bottom items in $\Delta \mathbf{P}$, or perform a weighted summation using the values of ${e_{i,j}}$ and 
$g_{i,j}(\mathbf{h}_i)$. These variations allow for more flexible and refined scenario predictions.

\begin{figure*}[t]
\centering
  \includegraphics[width=1.0\linewidth]{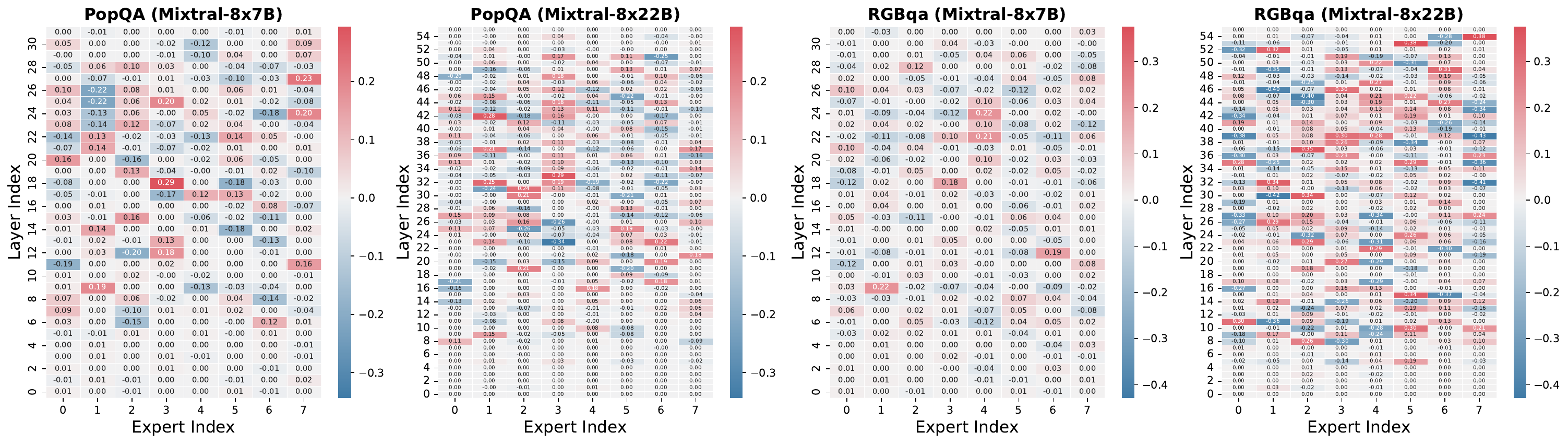}
  \caption{
  The visualization results of the cognizant expert. 
  Each value represents the contrastive activation probability of the expert, with deeper colors indicating the higher absolute value of activation probability. 
}
\vspace{-0.2cm}
 \label{fig:knowledge_expert}
\end{figure*}

\section{Inspecting Core Experts for RAG}
\label{sec:detecting_core_experts}
In this section, we employ CEAI to inspect expert activation and discover three types of core experts to address RAG challenges. 
In \S \ref{sec:knowledge_expert}, we discover cognizant experts that indicate the sufficiency of the model's internal knowledge, which can enhance RAG by avoiding unnecessary retrievals. 
\S \ref{sec:quality_expert} covers quality experts that can filter low-quality retrieved documents; 
In \S \ref{sec:retrieval_expert}, we show in-context experts that enhance the model's ability to utilize contextual information from retrieved documents.  
\subsection{Experimental Settings}

We use Mixtral-8x7B-instruct-v0.1 and Mixtral-8x22B-instruct-v0.1~\cite{jiang2024mixtral} in our experiments, as they stand out as widely used open-source MoE-based LLMs \cite{xue2024openmoe,bai2024patentgpt}. 
Our investigation is mainly conducted on question-answering datasets including PopQA~\cite{mallen-etal-2023-trust} and RGBqa \cite{chen2024benchmarking}, which are commonly utilized for RAG analysis.
We randomly select 1,000 samples from the PopQA and use the entire English subset of RGBqa with 300 samples.
For retrievers, both PopQA and RGBqa have released their retrieved question-related documents, thus we directly use these officially retrieved documents in our experiments.
We instruct the model to directly generate answers and apply CEAI at the first generated token.
Greedy decoding is used for all experiments for reproducibility.
Due to space constraints, we only introduce the main experimental results, and additional experimental results (e.g., other MoE-based LLMs) and more experimental details are provided in Appendix \ref{appendix:core_experts}.

\subsection{Cognizant Experts}
\label{sec:knowledge_expert}
Always retrieving external documents is not the optimal solution for RAG \cite{chen2024benchmarking}. 
The retrieved documents introduce additional inference cost, and low-quality retrieved documents can even mislead the LLMs \cite{shi2023large}. 
A more reasonable strategy is to retrieve only when the internal knowledge LLM is insufficient to answer the given question~\cite{asai2024selfrag}. 
We hypothesize that expert activation can indicate whether LLM's internal knowledge is sufficient. 
In this subsection, we discover these cognizant experts with CEAI.

\textbf{Setup.} 
We start by defining the contrastive scenarios for knowledge sufficiency.
For each sample in the RAG dataset, we input only the question to the LLM and get a response.
A positive scenario is that response contains the correct answer, which indicates LLM's knowledge is sufficient for this question. 
$D_\text{pos}$ consists of these answerable data.
Response without the correct answer is regarded as the negative scenario. $D_\text{neg}$ consists of unanswerable data.
The intuition behind our method is that certain experts specialize in specific types of knowledge and are frequently activated given knowledge-related questions. 
If the model lacks such knowledge and often answers incorrectly, the frequently activated experts in this scenario indicate model's knowledge is insufficient. We show the additional experimental details of cognizant experts in Appendix \ref{appendix:knowledge_setup}.

\begin{figure*}[t]
\centering
  \includegraphics[width=1.0\linewidth]{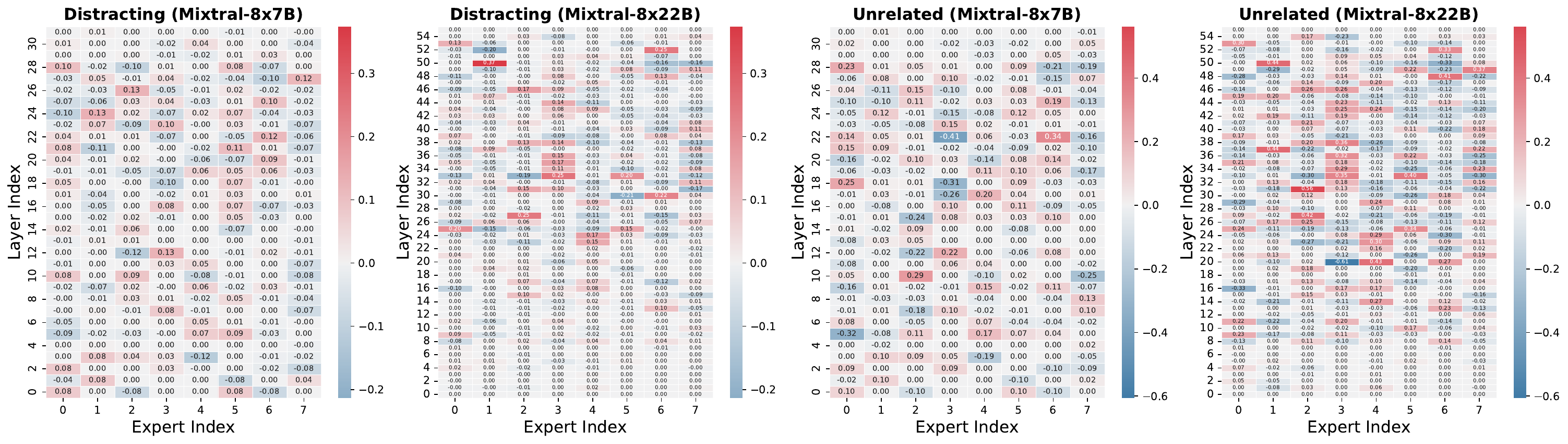}
  \caption{
  The visualization of the quality expert. 
 Each value represents the contrastive activation probability of the expert, with deeper colors indicating the higher absolute value of activation probability. 
}
\vspace{-0.2cm}
 \label{fig:quality_expert}
\end{figure*}

\textbf{Empirical Findings.} 
By applying CEAI to the $D_\text{pos}$ (model answers correctly) and $D_\text{neg}$ (model answers incorrectly), we get the contrastive activation probability for the cognizant expert and visualize it in Figure \ref{fig:knowledge_expert}.
We can observe that:
(1) 
there exists a clear distinction between the expert activation probabilities in both answerable and unanswerable scenarios.
 Such experts are widely present in all data and models, demonstrating the existence of cognizant experts.
(2) The cognizant experts differ across datasets. 
We speculate that each dataset requires different types of knowledge, which are possessed by different experts, resulting in diverse activation of cognizant experts. This may help uncover the knowledge distribution patterns hidden in various tasks and scenarios. 
(3) For the larger model Mixtral-8x22B, the value of contrastive activation probability is generally higher compared to  Mixtral-8x7B. 
One possible explanation is that the limited number of experts in smaller models forces each expert to share a wider range of knowledge and abilities during pre-training. 
Conversely, larger models have more experts, allowing for more distinct specialization.

\begin{table}[t!]
    \resizebox{\linewidth}{!}{
    \begin{tabular}{llcc}
    \toprule 
    \textbf{Model} & \textbf{Method}      & \textbf{PopQA} & \textbf{RGBqa}\\
        \midrule
   \multirow{3}{*}{\textbf{Mixtral-8x7B}} & Random Guess  & 42.01  & 40.28   \\
    & Knowledge (50-Shot) &  55.87 & 55.15  \\
    & Knowledge (Full-Set) &  \textbf{56.77} & \textbf{59.84}  \\
    \midrule
    \multirow{3}{*}{\textbf{Mixtral-8x22B}} & Random Guess  & 47.39  & 47.18 \\
    &Knowledge (50-Shot) & 64.78 &  73.31 \\
    & Knowledge (Full-Set) &  \textbf{65.83} & \textbf{75.40}  \\
    \bottomrule
    \end{tabular}
    }
    \caption{F1-Score of predicting the sufficiency of LLM's knowledge. Bold numbers are the best performance.}
    \vspace{-0.4cm}
    \label{tab:knowledge_expert}
\end{table}

\textbf{Analyses.}
 If our identified cognizant experts are indeed responsible for knowledge sufficiency, their activation should be able to predict whether the model can answer a question correctly.
To test this hypothesis, we calculate a scenario score using the identified cognizant experts, which can predict if the model's knowledge is sufficient to answer the given question. 
We utilize the full set and randomly selected 50-shot subset to identify and search for the best cognizant experts, comparing them with a random guess baseline. 
Due to the imbalance between knowledge insufficiency and sufficiency data, we choose the F1-Score as the evaluation metric.
 The results in Table~\ref{tab:knowledge_expert} show that, across all settings, cognizant experts outperform random guessing in predicting scenarios. 
  Using the full set to identify cognizant experts leads to the best performance, while 50-shot also achieves impressive performance despite being a small fraction of the full set.
    With only the 50-shot, the Mixtral-8x7B achieves an absolute improvement of 13.86 percent over the random guessing baseline on the PopQA dataset. 
  This highlights the strong generalization capability of cognizant experts. Overall, these findings support the existence of the cognizant expert and show its potential to avoid unnecessary retrieval to enhance the efficiency of RAG.

\subsection{Quality Experts}
\label{sec:quality_expert}
A reason why retrieved documents are not always beneficial is that low-quality retrieved documents can mislead LLMs~\cite{shi2023large}. 
This motivates us to explore whether the expert activation can evaluate the quality of documents. 

\textbf{Setup.} 
We define the positive scenario as contexts containing high-quality documents and the negative scenario as contexts containing low-quality documents.  
Following \citet{chen2024benchmarking}, high-quality documents are those that contain the correct answer, while low-quality documents do not.
Low-quality documents are further divided into two categories: Distracting documents, which are related to the question but lack the correct answer, and Unrelated documents, which are not related to the question at all.
We use RGBqa to construct the contrastive dataset as it offers retrieved documents with different qualities that satisfy our requirements. 
The construction process and additional details are presented in Appendix \ref{appendix:quality_setup}.

\begin{table}[t!]
    \resizebox{\linewidth}{!}{
    \begin{tabular}{llcc}
    \toprule 
    \textbf{Model} & \textbf{Method}      & \textbf{Distracting} & \textbf{Unrelated} \\
    \midrule
   \multirow{3}{*}{\textbf{Mixtral-8x7B}} &Random Guess& 50.07 & 50.07\\
    & Quality (50-Shot) & 57.48 & 76.64 \\
    & Quality (Full-Set) & \textbf{62.50} & \textbf{80.66} \\
    \midrule
    \multirow{3}{*}{\textbf{Mixtral-8x22B}} & Random Guess & 50.07 & 50.07 \\
    &Quality (50-Shot) & 71.28 & 90.92 \\
    &Quality (Full-Set) &  \textbf{75.33}  & \textbf{93.00} \\
    \bottomrule
    \end{tabular}
    }
    \caption{Accuracy of predicting the quality of retrieved documents. Bold numbers are the best performance.}
    \vspace{-0.4cm}
    \label{tab:quality_expert}
\end{table}

\begin{figure*}[t!]
    \centering
    \includegraphics[width=1.0\linewidth]{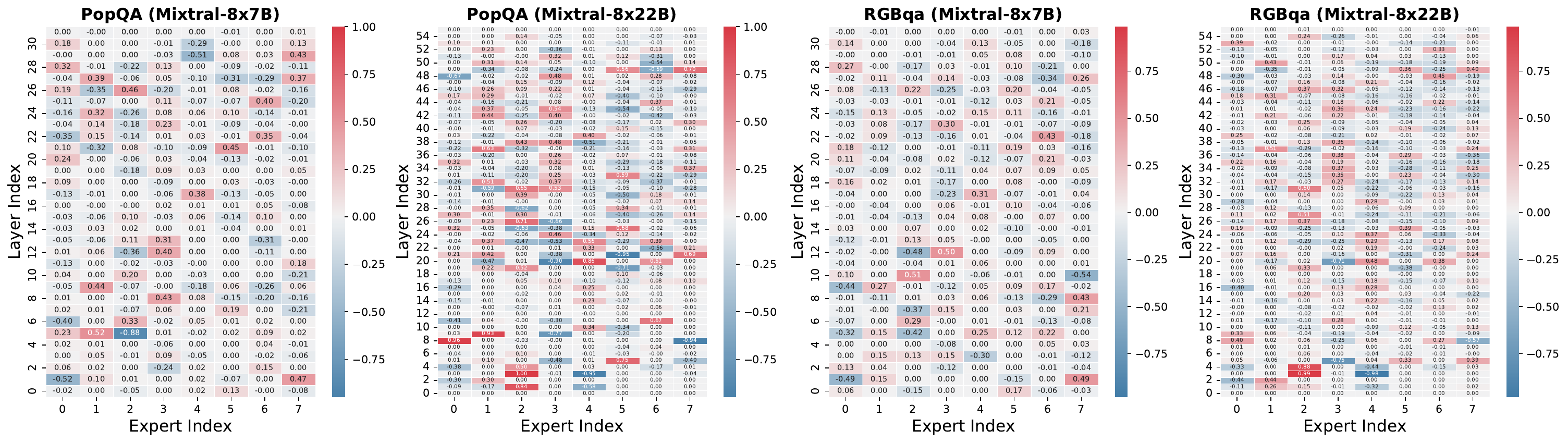} \vskip -0.8em
    \caption{
    The visualization results of the in-context expert. 
  Each value represents the contrastive activation probability of the expert, with deeper colors indicating the higher absolute value of activation probability. 
}
\vspace{-0.4cm}
    \label{fig:retrieval_experts}
\end{figure*}

\textbf{Empirical Findings.} 
Based on the results visualized in Figure \ref{fig:quality_expert}, we observe distinct differences in expert activation between contexts containing high-quality versus low-quality documents. 
Such differences increase as document quality decreases, further confirming the impact of retrieved document quality on expert activation. 
Moreover, the activation differences become more evident as the model scale increases, a trend similar to that observed with cognizant experts. 
These findings validate the existence of quality experts. 

\textbf{Analyses.}
To further investigate the effectiveness of quality experts, we use scenario scores to predict the quality of retrieved documents.
By combining $D_\text{pos}$ and $D_\text{neg}$, we construct the verification dataset that includes both high-quality and low-quality retrieved documents. Two datasets, Unrelated and Distracting, are created with different levels of low-quality documents. 
The experimental results in Table \ref{tab:quality_expert} demonstrate  the effectiveness of quality experts in distinguishing the quality of retrieved documents. 
We observe that the quality expert consistently outperforms random guessing across both model scales and datasets. On Mixtral-8x22B, full-set quality expert achieves a remarkable 93\% accuracy on the Unrelated dataset and a respectable 75.33\% on the Distracting dataset. The 50-shot also achieves impressive performance, only slightly lower than the full-set.
Similar to the observations made with cognizant experts, we find that the larger the difference in expert activation, the better the performance of the quality expert. 

\subsection{In-context Experts}
\label{sec:retrieval_expert}
The ability to leverage contextual information is crucial for RAG \cite{shi2023trusting}. Yet, LLMs may struggle to effectively extract information from context, even though the context includes high-quality retrieved documents. \cite{xie2023adaptive}. 
This motivates us to identify in-context experts that can enhance the LLM's ability to utilize context.

\textbf{Setup.} 
Intuitively, contrastive scenarios for using contextual information can be constructed by including or excluding the retrieved documents in context.
We view the data that consists of a question paired with retrieved documents as $D_\text{pos}$, whereas data that includes only the question without retrieved documents as $D_\text{neg}$.
To avoid potential bias from text length, we pad the texts in $D_\text{neg}$ to match the length of those in $D_\text{pos}$. 
We show in-context experts' detailed experimental setups for empirical findings and analyses in Appendix \ref{appendix:retrieval_setup}.

\textbf{Empirical Findings.} 
 Figure \ref{fig:retrieval_experts} reveals distinct differences in expert activation between scenarios where the context includes or excludes retrieved documents, validating the existence of in-context experts. 
Additionally, we find a subset of experts is frequently activated across all datasets, which suggests presence of universal in-context experts.

\begin{table}[t!]
\centering
    \resizebox{\linewidth}{!}{
    \begin{tabular}{llccc}
    \toprule 
    \textbf{Method} & \textbf{Expert Type}& \textbf{PopQA} & \textbf{RGBqa} \\
    \midrule
   \textbf{No Adjustment} & None & 50.70 & 89.33 \\
    \midrule
   \multirow{2}{*}{\textbf{Enhancement} $\uparrow$ }& Random & 50.03 & 88.00 & \\
   & In-context & 52.20 & 90.66 & \\
   \midrule
    \multirow{2}{*}{\textbf{Inhibition} $\downarrow$  } & Random & 47.60 & 83.00 & \\
    & In-context & 49.30 & 87.00 & \\
    \bottomrule
    \end{tabular}
    }
    \caption{Results of adjusting the activation of experts in Mixtral-8x7B.  $\uparrow$ indicates that high values are desirable. $\downarrow$ indicates that low values are desirable.}
    \vspace{-0.5cm}
    \label{tab:retrieval_expert}
\end{table}

\textbf{Analyses.}
We validate the effectiveness of in-context experts on RAG tasks.  
Guided by the contrastive activation probability $\Delta \mathbf{P}$ of in-context experts, we control the activation of experts during the model's forward pass, adjusting its ability to use the information in context.
To enhance this ability, we enforce the activation of experts with high $\Delta \mathbf{P}$ values and increase their weight, while preventing the activation of experts with low values.
Importantly, we do not increase the number of activated experts, each MoE module still activates the default number of experts. 
 From the results in Table \ref{tab:retrieval_expert}, we can observe that enhancing in-context experts improves the performance of RAG while inhibiting them leads to performance degradation. 
Given the complexity of the LLM's internal mechanisms, the activation of in-context experts can only play a limited role and cannot substantially control the model's behavior, which is within our expectations.  
Interestingly, random experts seem to be more effective in inhibiting model abilities compared to in-context experts, resulting in lower task performance. However, enhancing random experts does not lead to performance improvement, while enhancing in-context experts does. We hypothesize that this is because random experts do not intentionally control the target ability, but instead inhibit some general experts that are crucial to the general ability. 
In Appendix \ref{appendix:retrieval_setup}, we show the experimental details and further demonstrate that inhibiting general experts that highly activated in any scenario (but ignored by in-context experts due to contrast operation) causes more severe performance degradation than random experts.

\section{Application on Adaptive RAG}
\label{sec:application}
Recent research has increasingly focused on Adaptive RAG (ARAG), which reduces retrieval cost and enhances RAG performance by enabling retrieval only when necessary \cite{wang2023learning,asai2024selfrag}. 
Given the considerable role our identified experts fulfill in such scenarios, it is intuitive to use them to improve RAG.
In this section, we first introduce our method in \S \ref{sec:e_rag}, followed by the experimental setup in \S \ref{sec:application_setup}. 
 Finally, we show the effectiveness of our method in \S \ref{sec:application_results}. 

\subsection{Enhancing RAG via Expert Activation}
\label{sec:e_rag}
We introduce Expert-RAG, which utilizes three discovered experts to improve the effectiveness and adaptivity of RAG. 
Expert-RAG involves the following steps:
(1) \textbf{Knowledge Judgment}: Given a question, we first input it to LLM and collect the activation of experts during the forward phase. 
Then we utilize cognizant experts to calculate the scenario score to predict whether knowledge is sufficient based on the expert activation of the given question. 
Retrieval is only enabled when the cognizant experts predict that the model's internal knowledge is insufficient.
(2) \textbf{Quality Filter}: Once documents are retrieved, we input the question with retrieved documents to LLM and use the quality expert to predict the quality of documents. 
Only high-quality documents are used for further processing.
(3) \textbf{Retrieval Enhancement}: With high-quality retrieved documents in context, we adjust the in-context experts to boost the model's ability to use context.
Our method only requires a few data to identify the above experts, which is training-free and easy to implement.

\begin{table*}[t!]
    \resizebox{\linewidth}{!}{
    \begin{tabular}{lcccccccccccccccc}
    \toprule 
    \multirow{2}{*}{\textbf{Method}} & \multicolumn{3}{c}{\textbf{PubHealth}} & & \multicolumn{3}{c}{\textbf{PopQA}} & & \multicolumn{3}{c}{\textbf{$\text{RGBqa}_{mix}$}} & & \multicolumn{3}{c}{\textbf{BalanceQA}} \\
    \cline{2-4} \cline{6-8} \cline{10-12} \cline{14-16}
    & Acc $\uparrow$ & R-Score $\uparrow$ & R-Token $\downarrow$ & & Acc $\uparrow$ & R-Score $\uparrow$ & R-Token $\downarrow$ & & Acc $\uparrow$ & R-Score $\uparrow$ & R-Token $\downarrow$ & & Acc $\uparrow$ & R-Score $\uparrow$ & R-Token $\downarrow$  \\
    \midrule
    \textbf{No RAG} & 51.87 & 51.87 & 0 & & 34.00 & 34.00 & 0 &  & 36.67 & 37.67 & 0 & & 50.00 & 50.00 & 0 \\
    \textbf{Always RAG} & 54.71 & 48.12 & 514274 & & \textbf{50.70} & 64.60 & 519858 & & 58.50 & 62.33 & 344916 & & 50.00 & 50.00 & 212650  \\
    \textbf{Random RAG} & 53.69 & 50.65 & \textbf{263647} & & 43.00 & 52.30 & \textbf{253148} & & 47.66 & 48.83 & \textbf{171273} & & 49.25 & 49.00 & 107241   \\
    \midrule
    \textbf{Expert-RAG} \\
    w/ C   & 56.94 & 62.51 & 371506 & & 49.70 & \textbf{72.60} & 443202 &  & 59.33 & 66.00 & 329934 & & 57.75 & 64.50 & 116545  \\
    w/ C\&Q    & 57.24 & 63.02 & 369555 & & 49.50 & 72.20 & 439033  &  & \textbf{60.67} & 66.00 & 297062 & & \textbf{59.00} & \textbf{64.70} & \textbf{104507}  \\
    w/ C\&Q\&R  & \textbf{58.05} & \textbf{63.02} & 369555 && 49.90  & 72.20 & 439033 & & 58.50 & \textbf{66.88} & 297062 && 58.75 & \textbf{64.70} & \textbf{104507} \\
    \bottomrule
    \end{tabular}
    }
\caption{Overall experiment results on RAG datasets.
Bold numbers indicate the best-desired performance among RAG baselines. $\uparrow$ means higher is better and $\downarrow$ means lower is better.
Acc is the Accuracy (task performance); R-Token is the token number of retrieved documents (inference cost). R-Score is the retrieval score. 
For Expert-RAG, C is cognizant expert, Q is quality expert and R is in-context expert. w/ C\&Q\&R means using all three experts. }
\label{tab:main_results}
\end{table*}

\subsection{Experimental Setup}
\label{sec:application_setup}
We first highlight the limitations of the current ARAG evaluation and then describe our main experimental setup. Additional details about data composition, metrics, prompts and hyperparameters are provided in Appendix  \ref{appendix:RAG_setup}.

\textbf{RAG Evaluation.}
Most RAG datasets fail to effectively reflect the advantages of ARAG, as always retrieving documents often yields the best task performance on these data. 
Using only task performance as the evaluation metric cannot reflect the efficiency advantages of ARAG, and the retrieved documents can also be harmful in real-world applications.
To address these issues, we propose evaluation metrics and data recipes to make ARAG evaluation more comprehensive.

\textbf{Metric.} 
For task performance, we follow previous work \cite{shi2023trusting,asai2024selfrag} and use Accuracy (Acc) as the task performance metric for all datasets. We mark a prediction as correct if any substring of the prediction exactly matches any of the gold answers. 
Additionally, we use Retrieval Score (R-Score) to evaluate the necessity of retrieval and R-Token to evaluate the additional inference cost introduced by RAG. 
We consider the R-Score measures the accuracy between the model's actual retrieval requirements and the ARAG method's retrieval predictions.
R-Token represents the token length of the retrieved documents used for the generation, which relies solely on the retrieved documents and the tokenizer, allowing comparisons across devices and platforms. 
These two metrics allow us to evaluate the efficiency and effectiveness of ARAG comprehensively. 

\textbf{Dataset.}
We follow previous RAG work \cite{chen2024benchmarking,asai2024selfrag} and select commonly used QA datasets for evaluation, including PopQA \cite{mallen-etal-2023-trust}, RGBqa \cite{chen2024benchmarking}, and PubHealth \cite{asai2024selfrag}. 
To better demonstrate the effectiveness of ARAG, we designed a data recipe called BalanceQA.
BalanceQA consists of 50\% questions that the model can answer correctly and 50\% that it cannot without RAG.
For each question type, half is provided high-quality retrieved documents that make LLM answer correctly, while the other half is provided low-quality documents that lead to incorrect answers. 
This setup is designed to simulate real-world scenarios where retrieved documents may provide misleading information.  
With this recipe, no retrieval, always retrieval, and random retrieval methods would achieve an Acc and R-Score of 50\%, and an effective ARAG method would surpass 50\% for both metrics.  
Any data meeting the requirements can be used to construct this data recipe and improve the ARAG evaluation.
We use PubHealth, RGBqa, and PopQA to construct BalanceQA in our experiments. 

\textbf{Baselines.}
Our experiments are conducted on the Mixtral series. We present the results of the Mixtral-8x7B-instruct-v0.1 in \S \ref{sec:application_results}. 
We compare Expert-RAG with three baselines:
No RAG does not provide retrieved documents in context;
Always RAG consistently provides retrieved documents in the context; 
Random RAG randomly provides retrieved documents in the context with a probability of 50\%. 
Additionally, we also conduct experiments on other baselines, datasets and LLMs,  with detailed results presented in Appendix \ref{appendix:RAG_setup} for clearer illustration.

\subsection{Results}
\label{sec:application_results}
Table \ref{tab:main_results} presents the main results of our experiments.
We can observe that (1)  \textit{Always RAG} consistently improves task performance across nearly all datasets, except \textit{BalanceQA}.  However, the R-Score and R-Token reveal that there are many unnecessary retrievals required by Always RAG, which introduce additional costs. 
In contrast, our proposed \textit{Expert-RAG} consistently achieves competitive performance with reduced retrieved tokens compared to the \textit{Always RAG} across all datasets. We even outperform \textit{Always RAG} on \textit{RGBqa}, \textit{PubHealth}, and \textit{BalanceQA}. This is because we cannot guarantee that the retrieved documents always contain information beneficial to the current question. Always using retrieved documents may interfere LLM's internal knowledge, leading to performance degradation. 
(2) \textit{BalanceQA} introduces ``risk'' into retrieval, which prevents heuristic methods like \textit{No RAG} and \textit{Always RAG} from achieving optimal performance, thereby highlighting the advantages of ARAG. 
Our method achieves a 9\% improvement on the Acc score and a 14.7\% improvement on R-Score on BalanceQA, demonstrating its effectiveness as an ARAG method. 
(3) As the core of ARAG, the cognizant expert brings the most improvements across all datasets. The quality expert can mitigate negative impact of frequently retrieved low-quality documents, while in-context experts are beneficial when contextual information is reliable.
Overall, these results confirm the effectiveness of our metric and data recipe and show that it is practical to enhance RAG with expert activation.

\section{Related Work}
\paragraph{Mixture-of-Experts}
By replacing the dense FFN layer with dynamically activated experts \cite{jacobs1991adaptive}, MoE greatly enhances model performance without increasing the number of activated parameters and thus is widely used in LLMs \cite{shazeer2016outrageously,du2022glam,jiang2024mixtral,deepseekai2024deepseekv2}.
As core mechanisms of MoE, the routing network activates the appropriate experts based on different input representations and scenario demands, thereby influencing model behavior \cite{zhou2022mixture,chi2022representation}. This inspire us that expert activation can indicate scenario and adjust model behavior. 

\paragraph{Retrieval-augmented LLM}
In complex real-world applications, knowledge within LLMs' parameters \cite{brown2020language,touvron2023llama,jiang2024mixtral}  is usually insufficient or out-of-date, leading to hallucinations \cite{cao2020factual,10.1145/3571730,xu2024hallucination}. 
To mitigate this issue,  retrieval-augmented generation (RAG) enhances the LLM's input by retrieving query-relevant documents, offering external knowledge to improve the reliability of responses  \cite{guu2020retrieval,borgeaud2022improving,ren2023investigating}. 
However, RAG still faces many challenges \cite{gao2024retrievalaugmented,chen2024benchmarking}, such as the inference costs due to the lengthy retrieved documents \cite{xu2023retrieval}, information interference from low-quality retrieved documents \cite{shi2023large}, and the model's mistrust of the retrieved documents \cite{xie2023adaptive,yu-etal-2023-characterizing}. 
To address these problems, some works focus on improving retriever quality \cite{pan2024i3,ke2024bridging} or refining the retrieved documents  \cite{xie2023adaptive,wang2023learning}. Other works focus on training RAG-specific models \cite{lin2023ra}, which require high training costs. 
Unlike these methods that improve RAG from an external perspective, we investigate the impact of internal expert activation within MoE-based LLM in various RAG scenarios and provide a low-cost adaptive RAG solution.

\section{Conclusion}
In this paper, we explore the impact of expert activation within MoE-based LLMs in the context of RAG.
We introduce CEAI, a method that compares differences of expert activation frequency in contrasting scenarios to identify core experts responsible for specific scenarios. 
We identify three types of core experts for RAG: cognizant experts, quality experts, and in-context experts.
We further demonstrate how the activation of core experts can predict scenarios and enhance model behaviors. 
Building on these insights, we propose an expert-based adaptive RAG method and several methods for comprehensive ARAG evaluation.
Our experiments across multiple datasets confirmed the effectiveness of enhancing RAG via expert activation.

\section*{Limitation}
The limitations of this work are: (1) The focus of this work is the impact of expert activation in MoE-based LLM on RAG. Given that all activated experts can be regarded as a natural subnetwork, there may also be such a subnetwork in a dense network. However, unlike MoE models, dense models do not have naturally activated experts that allow us to find such subnetworks directly.  Since we focus on the expert activation in MoE-based LLM, the dense model is beyond the scope of this paper, and we leave this direction for future work.
(2) Our experiments are conducted on instruction-tuned MoE-based LLMs. We did not evaluate the base model without instruction fine-tuning or models specifically designed for RAG. Despite this, our experiments included models of various scales, and we identified the same core experts in Qwen models with different MoE architectures and training methods. This finding suggests that the three types of experts we discovered are universally present and that our approach is generalizable.

\section*{Acknowledgements}
The authors wish to thank the anonymous reviewers for their helpful comments. This work was partially funded by National Natural Science Foundation of China (No.62076069,62206057,61976056), Shanghai Rising-Star Program (23QA1400200), and Natural Science Foundation of Shanghai (23ZR1403500).

\bibliography{custom}

\begin{thebibliography}{36}
\providecommand{\natexlab}[1]{#1}

\bibitem[{Asai et~al.(2024)Asai, Wu, Wang, Sil, and Hajishirzi}]{asai2024selfrag}
Akari Asai, Zeqiu Wu, Yizhong Wang, Avirup Sil, and Hannaneh Hajishirzi. 2024.
\newblock \href {https://openreview.net/forum?id=hSyW5go0v8} {Self-{RAG}: Learning to retrieve, generate, and critique through self-reflection}.
\newblock In \emph{The Twelfth International Conference on Learning Representations}.

\bibitem[{Bai et~al.(2023)Bai, Bai, Chu, Cui, Dang, Deng, Fan, Ge, Han, Huang, Hui, Ji, Li, Lin, Lin, Liu, Liu, Lu, Lu, Ma, Men, Ren, Ren, Tan, Tan, Tu, Wang, Wang, Wang, Wu, Xu, Xu, Yang, Yang, Yang, Yang, Yao, Yu, Yuan, Yuan, Zhang, Zhang, Zhang, Zhang, Zhou, Zhou, Zhou, and Zhu}]{qwen}
Jinze Bai, Shuai Bai, Yunfei Chu, Zeyu Cui, Kai Dang, Xiaodong Deng, Yang Fan, Wenbin Ge, Yu~Han, Fei Huang, Binyuan Hui, Luo Ji, Mei Li, Junyang Lin, Runji Lin, Dayiheng Liu, Gao Liu, Chengqiang Lu, Keming Lu, Jianxin Ma, Rui Men, Xingzhang Ren, Xuancheng Ren, Chuanqi Tan, Sinan Tan, Jianhong Tu, Peng Wang, Shijie Wang, Wei Wang, Shengguang Wu, Benfeng Xu, Jin Xu, An~Yang, Hao Yang, Jian Yang, Shusheng Yang, Yang Yao, Bowen Yu, Hongyi Yuan, Zheng Yuan, Jianwei Zhang, Xingxuan Zhang, Yichang Zhang, Zhenru Zhang, Chang Zhou, Jingren Zhou, Xiaohuan Zhou, and Tianhang Zhu. 2023.
\newblock Qwen technical report.
\newblock \emph{arXiv preprint arXiv:2309.16609}.

\bibitem[{Bai et~al.(2024)Bai, Zhang, Chen, Cai, Zhong, Fang, Fang, Sun, Wang, Zhou et~al.}]{bai2024patentgpt}
Zilong Bai, Ruiji Zhang, Linqing Chen, Qijun Cai, Yuan Zhong, Cong Wang~Yan Fang, Jie Fang, Jing Sun, Weikuan Wang, Lizhi Zhou, et~al. 2024.
\newblock Patentgpt: A large language model for intellectual property.
\newblock \emph{arXiv preprint arXiv:2404.18255}.

\bibitem[{Borgeaud et~al.(2022)Borgeaud, Mensch, Hoffmann, Cai, Rutherford, Millican, Van Den~Driessche, Lespiau, Damoc, Clark et~al.}]{borgeaud2022improving}
Sebastian Borgeaud, Arthur Mensch, Jordan Hoffmann, Trevor Cai, Eliza Rutherford, Katie Millican, George~Bm Van Den~Driessche, Jean-Baptiste Lespiau, Bogdan Damoc, Aidan Clark, et~al. 2022.
\newblock Improving language models by retrieving from trillions of tokens.
\newblock In \emph{International conference on machine learning}, pages 2206--2240. PMLR.

\bibitem[{Brown et~al.(2020)Brown, Mann, Ryder, Subbiah, Kaplan, Dhariwal, Neelakantan, Shyam, Sastry, Askell et~al.}]{brown2020language}
Tom Brown, Benjamin Mann, Nick Ryder, Melanie Subbiah, Jared~D Kaplan, Prafulla Dhariwal, Arvind Neelakantan, Pranav Shyam, Girish Sastry, Amanda Askell, et~al. 2020.
\newblock Language models are few-shot learners.
\newblock \emph{Advances in neural information processing systems}, 33:1877--1901.

\bibitem[{Cao et~al.(2020)Cao, Dong, Wu, and Cheung}]{cao2020factual}
Meng Cao, Yue Dong, Jiapeng Wu, and Jackie Chi~Kit Cheung. 2020.
\newblock Factual error correction for abstractive summarization models.
\newblock In \emph{Proceedings of the 2020 Conference on Empirical Methods in Natural Language Processing (EMNLP)}, pages 6251--6258.

\bibitem[{Chen et~al.(2024)Chen, Lin, Han, and Sun}]{chen2024benchmarking}
Jiawei Chen, Hongyu Lin, Xianpei Han, and Le~Sun. 2024.
\newblock Benchmarking large language models in retrieval-augmented generation.
\newblock In \emph{Proceedings of the AAAI Conference on Artificial Intelligence}, pages 17754--17762.

\bibitem[{Chi et~al.(2022)Chi, Dong, Huang, Dai, Ma, Patra, Singhal, Bajaj, Song, Mao et~al.}]{chi2022representation}
Zewen Chi, Li~Dong, Shaohan Huang, Damai Dai, Shuming Ma, Barun Patra, Saksham Singhal, Payal Bajaj, Xia Song, Xian-Ling Mao, et~al. 2022.
\newblock On the representation collapse of sparse mixture of experts.
\newblock \emph{Advances in Neural Information Processing Systems}, 35:34600--34613.

\bibitem[{Chowdhery et~al.(2023)Chowdhery, Narang, Devlin, Bosma, Mishra, Roberts, Barham, Chung, Sutton, Gehrmann et~al.}]{chowdhery2023palm}
Aakanksha Chowdhery, Sharan Narang, Jacob Devlin, Maarten Bosma, Gaurav Mishra, Adam Roberts, Paul Barham, Hyung~Won Chung, Charles Sutton, Sebastian Gehrmann, et~al. 2023.
\newblock Palm: Scaling language modeling with pathways.
\newblock \emph{Journal of Machine Learning Research}, 24(240):1--113.

\bibitem[{DeepSeek-AI et~al.(2024)DeepSeek-AI, Liu, Feng, Wang, Wang, Liu, Zhao, Dengr, Ruan, Dai, Guo, Yang, Chen, Ji, Li, Lin, Luo, Hao, Chen, Li, Zhang, Xu, Yang, Zhang, Ding, Xin, Gao, Li, Qu, Cai, Liang, Guo, Ni, Li, Chen, Yuan, Qiu, Song, Dong, Gao, Guan, Wang, Zhang, Xu, Xia, Zhao, Zhang, Li, Wang, Zhang, Zhang, Tang, Li, Tian, Huang, Wang, Zhang, Zhu, Chen, Du, Chen, Jin, Ge, Pan, Xu, Chen, Li, Lu, Zhou, Chen, Wu, Ye, Ma, Wang, Zhou, Yu, Zhou, Zheng, Wang, Pei, Yuan, Sun, Xiao, Zeng, An, Liu, Liang, Gao, Zhang, Li, Jin, Wang, Bi, Liu, Wang, Shen, Chen, Chen, Nie, Sun, Wang, Liu, Xie, Yu, Song, Zhou, Yang, Lu, Su, Wu, Li, Wei, Zhu, Xu, Huang, Li, Zhao, Sun, Li, Wang, Zheng, Zhang, Xiong, Zhao, He, Tang, Piao, Dong, Tan, Liu, Wang, Guo, Zhu, Wang, Zou, Zha, Ma, Yan, You, Liu, Ren, Ren, Sha, Fu, Huang, Zhang, Xie, Hao, Shao, Wen, Xu, Zhang, Li, Wang, Gu, Li, and Xie}]{deepseekai2024deepseekv2}
DeepSeek-AI, Aixin Liu, Bei Feng, Bin Wang, Bingxuan Wang, Bo~Liu, Chenggang Zhao, Chengqi Dengr, Chong Ruan, Damai Dai, Daya Guo, Dejian Yang, Deli Chen, Dongjie Ji, Erhang Li, Fangyun Lin, Fuli Luo, Guangbo Hao, Guanting Chen, Guowei Li, H.~Zhang, Hanwei Xu, Hao Yang, Haowei Zhang, Honghui Ding, Huajian Xin, Huazuo Gao, Hui Li, Hui Qu, J.~L. Cai, Jian Liang, Jianzhong Guo, Jiaqi Ni, Jiashi Li, Jin Chen, Jingyang Yuan, Junjie Qiu, Junxiao Song, Kai Dong, Kaige Gao, Kang Guan, Lean Wang, Lecong Zhang, Lei Xu, Leyi Xia, Liang Zhao, Liyue Zhang, Meng Li, Miaojun Wang, Mingchuan Zhang, Minghua Zhang, Minghui Tang, Mingming Li, Ning Tian, Panpan Huang, Peiyi Wang, Peng Zhang, Qihao Zhu, Qinyu Chen, Qiushi Du, R.~J. Chen, R.~L. Jin, Ruiqi Ge, Ruizhe Pan, Runxin Xu, Ruyi Chen, S.~S. Li, Shanghao Lu, Shangyan Zhou, Shanhuang Chen, Shaoqing Wu, Shengfeng Ye, Shirong Ma, Shiyu Wang, Shuang Zhou, Shuiping Yu, Shunfeng Zhou, Size Zheng, T.~Wang, Tian Pei, Tian Yuan, Tianyu Sun, W.~L. Xiao, Wangding Zeng, Wei An, Wen
  Liu, Wenfeng Liang, Wenjun Gao, Wentao Zhang, X.~Q. Li, Xiangyue Jin, Xianzu Wang, Xiao Bi, Xiaodong Liu, Xiaohan Wang, Xiaojin Shen, Xiaokang Chen, Xiaosha Chen, Xiaotao Nie, Xiaowen Sun, Xiaoxiang Wang, Xin Liu, Xin Xie, Xingkai Yu, Xinnan Song, Xinyi Zhou, Xinyu Yang, Xuan Lu, Xuecheng Su, Y.~Wu, Y.~K. Li, Y.~X. Wei, Y.~X. Zhu, Yanhong Xu, Yanping Huang, Yao Li, Yao Zhao, Yaofeng Sun, Yaohui Li, Yaohui Wang, Yi~Zheng, Yichao Zhang, Yiliang Xiong, Yilong Zhao, Ying He, Ying Tang, Yishi Piao, Yixin Dong, Yixuan Tan, Yiyuan Liu, Yongji Wang, Yongqiang Guo, Yuchen Zhu, Yuduan Wang, Yuheng Zou, Yukun Zha, Yunxian Ma, Yuting Yan, Yuxiang You, Yuxuan Liu, Z.~Z. Ren, Zehui Ren, Zhangli Sha, Zhe Fu, Zhen Huang, Zhen Zhang, Zhenda Xie, Zhewen Hao, Zhihong Shao, Zhiniu Wen, Zhipeng Xu, Zhongyu Zhang, Zhuoshu Li, Zihan Wang, Zihui Gu, Zilin Li, and Ziwei Xie. 2024.
\newblock \href {https://arxiv.org/abs/2405.04434} {Deepseek-v2: A strong, economical, and efficient mixture-of-experts language model}.
\newblock \emph{Preprint}, arXiv:2405.04434.

\bibitem[{Ding et~al.(2024)Ding, Fan, Ning, Wang, Li, Yin, Chua, and Li}]{ding2024survey}
Yujuan Ding, Wenqi Fan, Liangbo Ning, Shijie Wang, Hengyun Li, Dawei Yin, Tat-Seng Chua, and Qing Li. 2024.
\newblock A survey on rag meets llms: Towards retrieval-augmented large language models.
\newblock \emph{arXiv preprint arXiv:2405.06211}.

\bibitem[{Du et~al.(2022)Du, Huang, Dai, Tong, Lepikhin, Xu, Krikun, Zhou, Yu, Firat et~al.}]{du2022glam}
Nan Du, Yanping Huang, Andrew~M Dai, Simon Tong, Dmitry Lepikhin, Yuanzhong Xu, Maxim Krikun, Yanqi Zhou, Adams~Wei Yu, Orhan Firat, et~al. 2022.
\newblock Glam: Efficient scaling of language models with mixture-of-experts.
\newblock In \emph{International Conference on Machine Learning}, pages 5547--5569. PMLR.

\bibitem[{Gao et~al.(2024)Gao, Xiong, Gao, Jia, Pan, Bi, Dai, Sun, Wang, and Wang}]{gao2024retrievalaugmented}
Yunfan Gao, Yun Xiong, Xinyu Gao, Kangxiang Jia, Jinliu Pan, Yuxi Bi, Yi~Dai, Jiawei Sun, Meng Wang, and Haofen Wang. 2024.
\newblock \href {https://arxiv.org/abs/2312.10997} {Retrieval-augmented generation for large language models: A survey}.
\newblock \emph{Preprint}, arXiv:2312.10997.

\bibitem[{Guu et~al.(2020)Guu, Lee, Tung, Pasupat, and Chang}]{guu2020retrieval}
Kelvin Guu, Kenton Lee, Zora Tung, Panupong Pasupat, and Mingwei Chang. 2020.
\newblock Retrieval augmented language model pre-training.
\newblock In \emph{International conference on machine learning}, pages 3929--3938. PMLR.

\bibitem[{Jacobs et~al.(1991)Jacobs, Jordan, Nowlan, and Hinton}]{jacobs1991adaptive}
Robert~A Jacobs, Michael~I Jordan, Steven~J Nowlan, and Geoffrey~E Hinton. 1991.
\newblock Adaptive mixtures of local experts.
\newblock \emph{Neural computation}, 3(1):79--87.

\bibitem[{Ji et~al.(2023)Ji, Lee, Frieske, Yu, Su, Xu, Ishii, Bang, Madotto, and Fung}]{10.1145/3571730}
Ziwei Ji, Nayeon Lee, Rita Frieske, Tiezheng Yu, Dan Su, Yan Xu, Etsuko Ishii, Ye~Jin Bang, Andrea Madotto, and Pascale Fung. 2023.
\newblock \href {https://doi.org/10.1145/3571730} {Survey of hallucination in natural language generation}.
\newblock 55(12).

\bibitem[{Jiang et~al.(2024)Jiang, Sablayrolles, Roux, Mensch, Savary, Bamford, Chaplot, Casas, Hanna, Bressand et~al.}]{jiang2024mixtral}
Albert~Q Jiang, Alexandre Sablayrolles, Antoine Roux, Arthur Mensch, Blanche Savary, Chris Bamford, Devendra~Singh Chaplot, Diego de~las Casas, Emma~Bou Hanna, Florian Bressand, et~al. 2024.
\newblock Mixtral of experts.
\newblock \emph{arXiv preprint arXiv:2401.04088}.

\bibitem[{Kang and Liu(2023)}]{kang2023deficiency}
Haoqiang Kang and Xiao-Yang Liu. 2023.
\newblock Deficiency of large language models in finance: An empirical examination of hallucination.
\newblock \emph{arXiv preprint arXiv:2311.15548}.

\bibitem[{Ke et~al.(2024)Ke, Kong, Li, Zhang, Mei, and Bendersky}]{ke2024bridging}
Zixuan Ke, Weize Kong, Cheng Li, Mingyang Zhang, Qiaozhu Mei, and Michael Bendersky. 2024.
\newblock \href {https://arxiv.org/abs/2401.06954} {Bridging the preference gap between retrievers and llms}.
\newblock \emph{Preprint}, arXiv:2401.06954.

\bibitem[{Lewis et~al.(2020)Lewis, Perez, Piktus, Petroni, Karpukhin, Goyal, K{\"u}ttler, Lewis, Yih, Rockt{\"a}schel et~al.}]{lewis2020retrieval}
Patrick Lewis, Ethan Perez, Aleksandra Piktus, Fabio Petroni, Vladimir Karpukhin, Naman Goyal, Heinrich K{\"u}ttler, Mike Lewis, Wen-tau Yih, Tim Rockt{\"a}schel, et~al. 2020.
\newblock Retrieval-augmented generation for knowledge-intensive nlp tasks.
\newblock \emph{Advances in Neural Information Processing Systems}, 33:9459--9474.

\bibitem[{Lin et~al.(2023)Lin, Chen, Chen, Shi, Lomeli, James, Rodriguez, Kahn, Szilvasy, Lewis et~al.}]{lin2023ra}
Xi~Victoria Lin, Xilun Chen, Mingda Chen, Weijia Shi, Maria Lomeli, Richard James, Pedro Rodriguez, Jacob Kahn, Gergely Szilvasy, Mike Lewis, et~al. 2023.
\newblock Ra-dit: Retrieval-augmented dual instruction tuning.
\newblock In \emph{The Twelfth International Conference on Learning Representations}.

\bibitem[{Lozano et~al.(2023)Lozano, Fleming, Chiang, and Shah}]{lozano2023clinfoai}
Alejandro Lozano, Scott~L Fleming, Chia-Chun Chiang, and Nigam Shah. 2023.
\newblock \href {https://arxiv.org/abs/2310.16146} {Clinfo.ai: An open-source retrieval-augmented large language model system for answering medical questions using scientific literature}.
\newblock \emph{Preprint}, arXiv:2310.16146.

\bibitem[{Mallen et~al.(2023)Mallen, Asai, Zhong, Das, Khashabi, and Hajishirzi}]{mallen-etal-2023-trust}
Alex Mallen, Akari Asai, Victor Zhong, Rajarshi Das, Daniel Khashabi, and Hannaneh Hajishirzi. 2023.
\newblock \href {https://doi.org/10.18653/v1/2023.acl-long.546} {When not to trust language models: Investigating effectiveness of parametric and non-parametric memories}.
\newblock In \emph{Proceedings of the 61st Annual Meeting of the Association for Computational Linguistics (Volume 1: Long Papers)}, pages 9802--9822, Toronto, Canada. Association for Computational Linguistics.

\bibitem[{Pan et~al.(2024)Pan, Li, Wang, Fei, Song, Ji, Lin, Liu, Chua, and Tang}]{pan2024i3}
Kaihang Pan, Juncheng Li, Wenjie Wang, Hao Fei, Hongye Song, Wei Ji, Jun Lin, Xiaozhong Liu, Tat-Seng Chua, and Siliang Tang. 2024.
\newblock \href {https://arxiv.org/abs/2308.10025} {I3: Intent-introspective retrieval conditioned on instructions}.
\newblock \emph{Preprint}, arXiv:2308.10025.

\bibitem[{Ren et~al.(2023)Ren, Wang, Qu, Zhao, Liu, Tian, Wu, Wen, and Wang}]{ren2023investigating}
Ruiyang Ren, Yuhao Wang, Yingqi Qu, Wayne~Xin Zhao, Jing Liu, Hao Tian, Hua Wu, Ji-Rong Wen, and Haifeng Wang. 2023.
\newblock Investigating the factual knowledge boundary of large language models with retrieval augmentation.
\newblock \emph{arXiv preprint arXiv:2307.11019}.

\bibitem[{Shazeer et~al.(2016)Shazeer, Mirhoseini, Maziarz, Davis, Le, Hinton, and Dean}]{shazeer2016outrageously}
Noam Shazeer, Azalia Mirhoseini, Krzysztof Maziarz, Andy Davis, Quoc Le, Geoffrey Hinton, and Jeff Dean. 2016.
\newblock Outrageously large neural networks: The sparsely-gated mixture-of-experts layer.
\newblock In \emph{International Conference on Learning Representations}.

\bibitem[{Shi et~al.(2023{\natexlab{a}})Shi, Chen, Misra, Scales, Dohan, Chi, Sch{\"a}rli, and Zhou}]{shi2023large}
Freda Shi, Xinyun Chen, Kanishka Misra, Nathan Scales, David Dohan, Ed~H Chi, Nathanael Sch{\"a}rli, and Denny Zhou. 2023{\natexlab{a}}.
\newblock Large language models can be easily distracted by irrelevant context.
\newblock In \emph{International Conference on Machine Learning}, pages 31210--31227. PMLR.

\bibitem[{Shi et~al.(2023{\natexlab{b}})Shi, Han, Lewis, Tsvetkov, Zettlemoyer, and Yih}]{shi2023trusting}
Weijia Shi, Xiaochuang Han, Mike Lewis, Yulia Tsvetkov, Luke Zettlemoyer, and Scott Wen-tau Yih. 2023{\natexlab{b}}.
\newblock Trusting your evidence: Hallucinate less with context-aware decoding.
\newblock \emph{arXiv preprint arXiv:2305.14739}.

\bibitem[{Touvron et~al.(2023)Touvron, Lavril, Izacard, Martinet, Lachaux, Lacroix, Rozi{\`e}re, Goyal, Hambro, Azhar et~al.}]{touvron2023llama}
Hugo Touvron, Thibaut Lavril, Gautier Izacard, Xavier Martinet, Marie-Anne Lachaux, Timoth{\'e}e Lacroix, Baptiste Rozi{\`e}re, Naman Goyal, Eric Hambro, Faisal Azhar, et~al. 2023.
\newblock Llama: Open and efficient foundation language models.
\newblock \emph{arXiv preprint arXiv:2302.13971}.

\bibitem[{Wang et~al.(2023)Wang, Araki, Jiang, Parvez, and Neubig}]{wang2023learning}
Zhiruo Wang, Jun Araki, Zhengbao Jiang, Md~Rizwan Parvez, and Graham Neubig. 2023.
\newblock Learning to filter context for retrieval-augmented generation.
\newblock \emph{arXiv preprint arXiv:2311.08377}.

\bibitem[{Xie et~al.(2023)Xie, Zhang, Chen, Lou, and Su}]{xie2023adaptive}
Jian Xie, Kai Zhang, Jiangjie Chen, Renze Lou, and Yu~Su. 2023.
\newblock Adaptive chameleon or stubborn sloth: Revealing the behavior of large language models in knowledge conflicts.
\newblock In \emph{The Twelfth International Conference on Learning Representations}.

\bibitem[{Xu et~al.(2023)Xu, Ping, Wu, McAfee, Zhu, Liu, Subramanian, Bakhturina, Shoeybi, and Catanzaro}]{xu2023retrieval}
Peng Xu, Wei Ping, Xianchao Wu, Lawrence McAfee, Chen Zhu, Zihan Liu, Sandeep Subramanian, Evelina Bakhturina, Mohammad Shoeybi, and Bryan Catanzaro. 2023.
\newblock Retrieval meets long context large language models.
\newblock In \emph{The Twelfth International Conference on Learning Representations}.

\bibitem[{Xu et~al.(2024)Xu, Jain, and Kankanhalli}]{xu2024hallucination}
Ziwei Xu, Sanjay Jain, and Mohan Kankanhalli. 2024.
\newblock \href {https://arxiv.org/abs/2401.11817} {Hallucination is inevitable: An innate limitation of large language models}.
\newblock \emph{Preprint}, arXiv:2401.11817.

\bibitem[{Xue et~al.(2024)Xue, Zheng, Fu, Ni, Zheng, Zhou, and You}]{xue2024openmoe}
Fuzhao Xue, Zian Zheng, Yao Fu, Jinjie Ni, Zangwei Zheng, Wangchunshu Zhou, and Yang You. 2024.
\newblock Openmoe: An early effort on open mixture-of-experts language models.
\newblock \emph{arXiv preprint arXiv:2402.01739}.

\bibitem[{Yu et~al.(2023)Yu, Merullo, and Pavlick}]{yu-etal-2023-characterizing}
Qinan Yu, Jack Merullo, and Ellie Pavlick. 2023.
\newblock \href {https://doi.org/10.18653/v1/2023.emnlp-main.615} {Characterizing mechanisms for factual recall in language models}.
\newblock In \emph{Proceedings of the 2023 Conference on Empirical Methods in Natural Language Processing}, pages 9924--9959, Singapore. Association for Computational Linguistics.

\bibitem[{Zhou et~al.(2022)Zhou, Lei, Liu, Du, Huang, Zhao, Dai, Le, Laudon et~al.}]{zhou2022mixture}
Yanqi Zhou, Tao Lei, Hanxiao Liu, Nan Du, Yanping Huang, Vincent Zhao, Andrew~M Dai, Quoc~V Le, James Laudon, et~al. 2022.
\newblock Mixture-of-experts with expert choice routing.
\newblock \emph{Advances in Neural Information Processing Systems}, 35:7103--7114.

\end{thebibliography}

\appendix
\section{Additional Details of CEAI}
\label{appendix:CEAI}
In our experiments, CEAI is applied to the last token of the input sequence. This position is also the position for generating the first token. Implementing CEAI in this position avoids the computationally intensive generation process, thereby enhancing the efficiency of RAG systems. Our method leverages expert activations to make critical decisions during the retrieval process, such as whether to retrieve documents and how to utilize the retrieved information. 
Therefore, applying CEAI before the generating phase can avoid generating all the responses and improve efficiency.

\section{Additional Details of Detecting Core Experts}
\label{appendix:core_experts}
\subsection{General Experimental Settings}
 This section details the experimental settings for identifying core experts on different scenarios.
\paragraph{Models}  We use Mixtral-8x7B-instruct-v0.1, Mixtral-8x22B-instruct-v0.1~\cite{jiang2024mixtral}, and Qwen1.5-MoE-A2.7B-Chat \cite{qwen} in our experiments.
Mixtral-8x7B consists of 32 layers; each layer contains 8 experts, and by default, only two experts are activated per layer.
Mixtral-8x22B consists of 56 layers, maintaining the same configuration of 8 experts per layer and activating two experts by default. 
QWEN1.5-A2.7B comprises 24 layers, with each layer having 4 always-activated shared experts and 60 dynamic experts, out of which 4 are activated by default. For all experiments, we utilized the default configurations of the above models. During generation, we employed greedy decoding to improve reproducibility. 
These models were implemented using the Huggingface framework. All experiments were conducted on  8xNVIDIA-A100-80GB.

\paragraph{Retrieved Documents} For PopQA\footnote{https://github.com/AlexTMallen/adaptive-retrieval} and RGBqa\footnote{https://github.com/chen700564/RGB}, we use the officially provided retrieved documents. 
In PopQA, each question was accompanied by five retrieved documents in the context.
In RGBqa, the type and quantity of documents vary according to the specific experimental conditions we detailed in relevant sections.
For PubHealth, following \cite{asai2024selfrag}, we use Contriever-MS MARCO to retrieve the top five documents from Wikipedia, utilizing the official Wikipedia embeddings based on the 2018 English Wikipedia.

\paragraph{Prompts} Following previous work \cite{asai2024selfrag}, we instruct the LLMs to directly generate answers. This also benefits CEAI, as CEAI is applied at the position of the first generated token. 
The prompts for different tasks are shown in Table \ref{tab:prompts} and are used consistently across all experiments.
In scenarios where contexts do not contain retrieved documents, such as identifying cognizant experts,  we filled the \{retrieved document\} placeholder with ``No paragraph available.'' Additionally, after applying the chat template to the prompt, we appended "The answer is: " to further ensure that the first generated token is the answer.

\paragraph{Evaluation Metric} For the RAG task, we use Exact Match as the performance metric. We lowercase both the LLM outputs and the answers and check if the correct answer exactly matches any part of the model's output. We use the accuracy between the core expert's prediction and the real scenario to measure the scenario prediction. 

\paragraph{Variant of Scenario Score} As discussed in \S \ref{sec:activation_as_classifier}, we explore various methods to compute scenario scores, treating these methods as hyperparameters for enhanced scenario prediction accuracy. We search these hyperparameters for all experiments and report the best performance.  For each dataset, we select and retain only the Top-k and Bottom-k experts based on contrastive activation probability $\Delta \mathbf{P}$.  These experts are identified as core experts for specific scenarios.  Furthermore, we can use the contrastive probability of core expert $P_{e_{i,j}}$ and the expert activation probability of the current input $g_{i,j}$ as the weights to calculate the score. 
We can also utilize the activation without any weights to cacluate score, such as $\sum_{i=1}^{L} \sum_{j=1}^{N} \mathbb{I}_1{(P_{e_{i,j}})} \cdot \mathbb{I}_2(g_{i,j}(\mathbf{x}))$, where $\mathbb{I}_1{(P_{e_{i,j}})} \rightarrow \{0,1\}$ means $e_{i,j}$ is the selected core experts and $\mathbb{I}_2(g_{i,j}(x)) \rightarrow \{0,1\}$ means $e_{i,j}$ is activated for current input.

\subsection{Results of Qwen}
\label{sec:qwen}
\begin{figure*}[t!]
    \centering
    \includegraphics[width=1.0\linewidth]{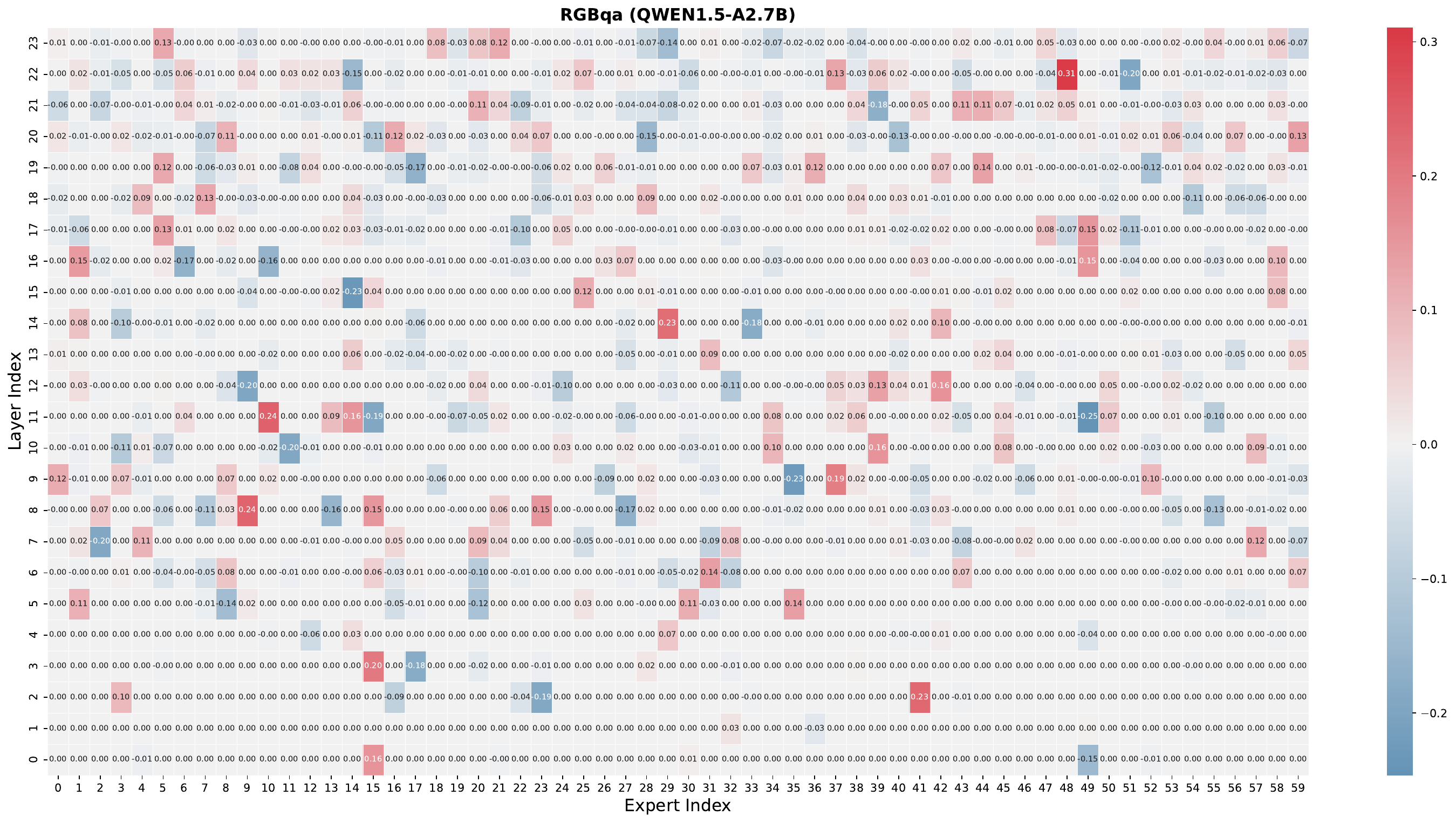}
    \caption{
    The visualization results of  QWEN1.5-MoE-A2.7B-Chat's cognizant expert. 
  Each value represents the activation probability of the expert in the corresponding scenario, with deeper colors indicating higher activation probability. 
}
    \label{fig:knowledge_experts_qwen_rgbqa}
\end{figure*}

\begin{figure*}[t!]
    \centering
    \includegraphics[width=1.0\linewidth]{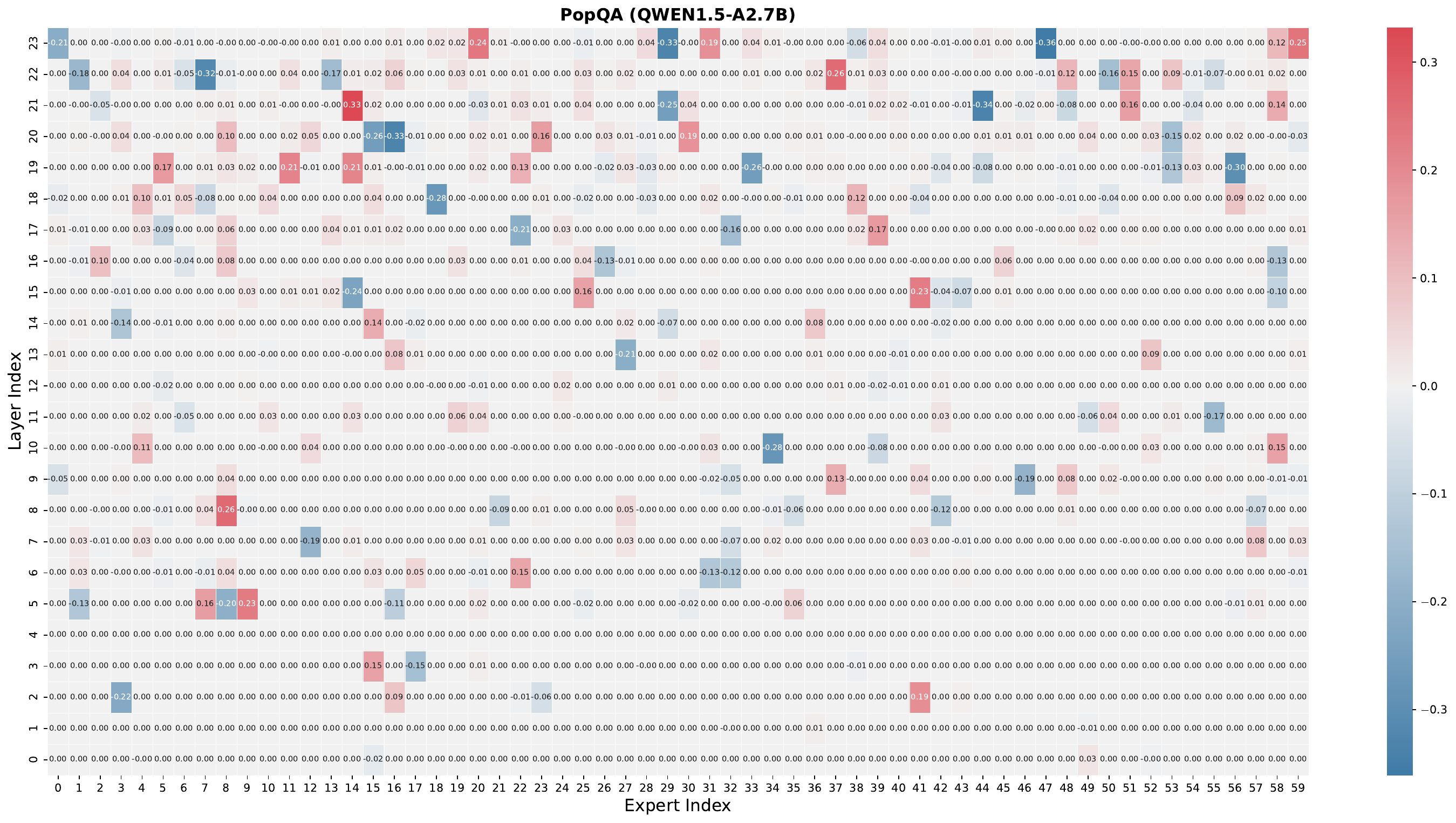}
    \caption{
    The visualization results of  QWEN1.5-MoE-A2.7B-Chat's cognizant expert. 
  Each value represents the activation probability of the expert in the corresponding scenario, with deeper colors indicating higher activation probability. 
}
    \label{fig:knowledge_experts_qwen_popqa}
\end{figure*}

\begin{figure*}[t!]
    \centering
    \includegraphics[width=1.0\linewidth]{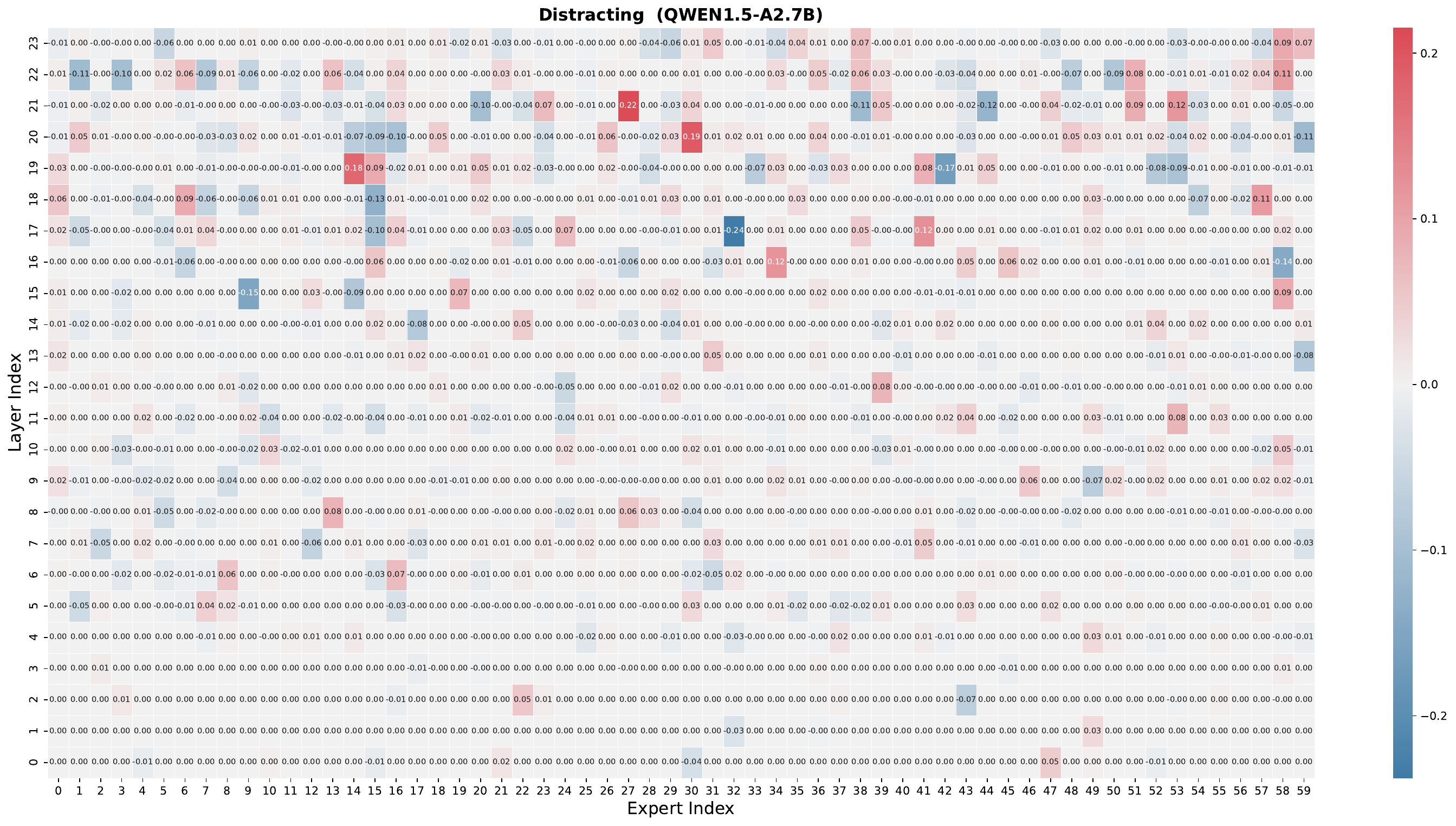}
    \caption{
    The visualization results of  QWEN1.5-MoE-A2.7B-Chat's quality expert. 
  Each value represents the activation probability of the expert in the corresponding scenario, with deeper colors indicating higher activation probability. 
}
    \label{fig:qwen_quality_expert_irrelevant}
\end{figure*}

\begin{figure*}[t!]
    \centering
    \includegraphics[width=1.0\linewidth]{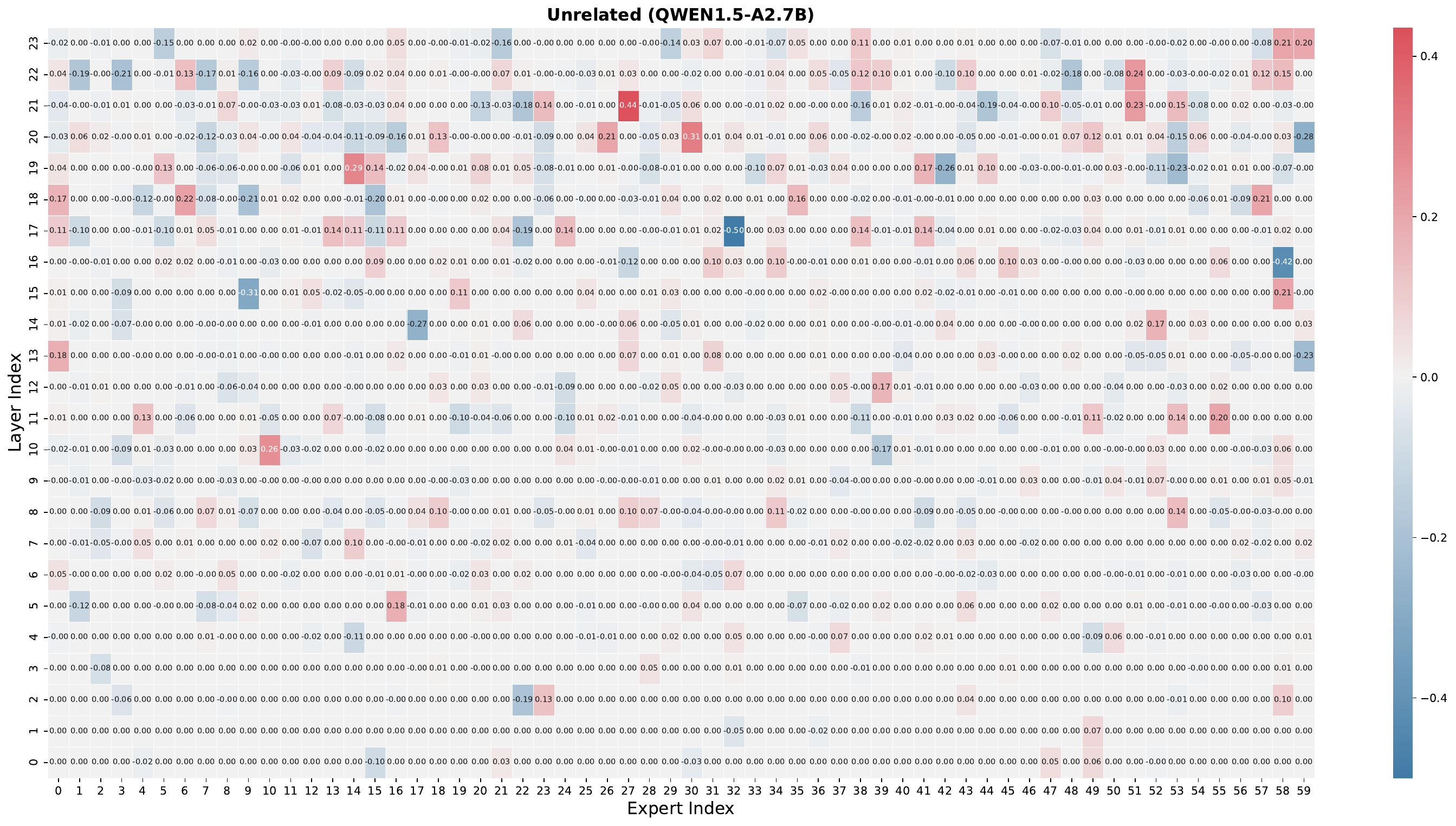}
    \caption{
    The visualization results of  QWEN1.5-MoE-A2.7B-Chat's quality expert. 
  Each value represents the activation probability of the expert in the corresponding scenario, with deeper colors indicating higher activation probability. 
}
    \label{fig:qwen_quality_expert_unrelated}
\end{figure*}

\begin{figure*}[t!]
    \centering
    \includegraphics[width=1.0\linewidth]{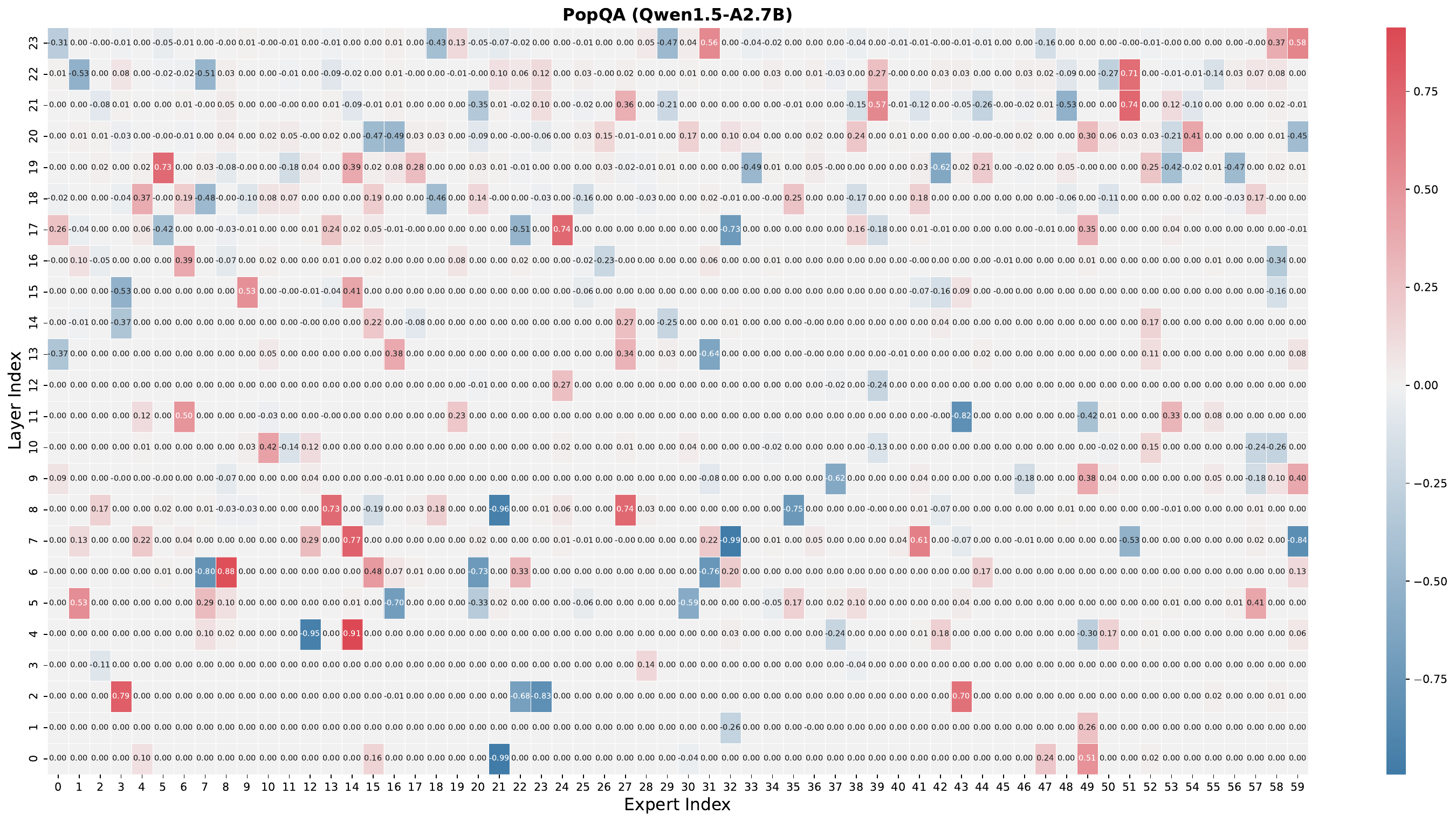}
    \caption{
    The visualization results of  QWEN1.5-MoE-A2.7B-Chat's in-context expert. 
  Each value represents the activation probability of the expert in the corresponding scenario, with deeper colors indicating higher activation probability. 
}
    \label{fig:qwen_retrieval_expert_popqa}
\end{figure*}

\begin{figure*}[t!]
    \centering
    \includegraphics[width=1.0\linewidth]{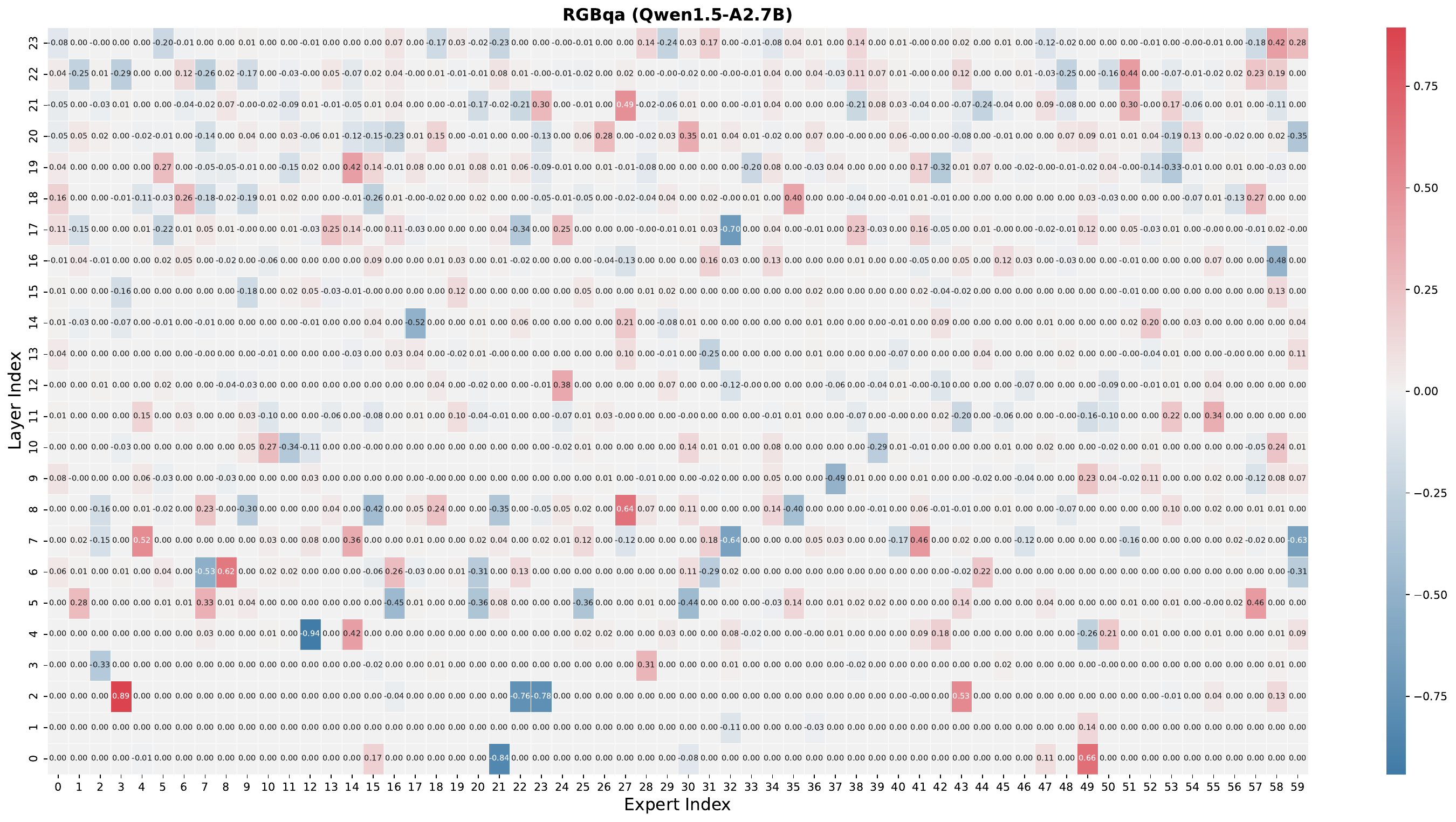}
    \caption{
    The visualization results of  QWEN1.5-MoE-A2.7B-Chat's in-context expert. 
  Each value represents the activation probability of the expert in the corresponding scenario, with deeper colors indicating higher activation probability. 
}
    \label{fig:qwen_retrieval_expert_RGBqa}
\end{figure*}

 As mentioned in \S \ref{sec:detecting_core_experts}, we also conduct experiments on Qwen1.5-MoE-A2.7B-Chat model. This model has fewer parameters but more experts than the Mixtral series, featuring 60 dynamically activated experts. Given the extensive number of experts, comprehensive visualization of the results within the main text is impractical. Therefore, we present the visualization of Qwen's core experts here. 
The visualization of cognizant expert activations is shown in  Figure \ref{fig:knowledge_experts_qwen_rgbqa}  and Figure \ref{fig:knowledge_experts_qwen_popqa}. 
Visualizations for quality experts are displayed in Figure \ref{fig:qwen_quality_expert_irrelevant} and Figure \ref{fig:qwen_quality_expert_unrelated}.
The in-context expert visualizations are also shown in Figure  \ref{fig:qwen_retrieval_expert_popqa} and Figure \ref{fig:qwen_retrieval_expert_RGBqa}. 

To assess the effectiveness of the various expert groups within the Qwen model, we also conduct experiments to evaluate whether these experts can be used to predict scenarios. The results are shown in Table \ref{tab:qwen}. 
Despite the large number of experts and distinct MoE architecture, 
the expert activation patterns in QWEN are also similar to the Mixtral series. Specifically, the following observations were made: there are different expert activations in contrastive scenarios, activation of core experts can be used to predict scenarios, and more evident difference leads to higher prediction accuracy. The phenomena described above have also been similarly observed among various core experts in the Mixtral series models. This observation underscores the ubiquity of core experts within models structured around MoE-based LLM, irrespective of their specific architectures. Furthermore, these findings imply a robust generalizability of our proposed method.

\begin{table}[t!]
    \centering
    \resizebox{\linewidth}{!}{
    \begin{tabular}{llcc}
    \toprule 
    \textbf{Model} & \textbf{Method}      & \textbf{PopQA} & \textbf{RGBqa}   \\
    \midrule
    \multirow{3}{*}{\textbf{Knowledge Expert}} & Random Guess & 31.59 & 23.11 \\
    & 50-shot & 50.96 & 37.94 \\
    & Full-Set & \textbf{52.25} & \textbf{44.23} \\
    \midrule
    \midrule
    \textbf{Model} & \textbf{Method}      & \textbf{Irrelevant} & \textbf{Unrelated}   \\
    \midrule
    \multirow{3}{*}{\textbf{Quality Expert}} & Random Guess & 50.70 & 50.70 \\
    & 50-Shot & 64.24 &  80.56 \\
    & Full Set &  \textbf{73.50}  &  \textbf{89.83} \\
    \bottomrule
    \end{tabular}
    }
    \caption{Results of Qwen1.5-MoE-A2.7B-Chat's scenarios prediction. }
    \label{tab:qwen}
\end{table}

\begin{table*}[t]
\centering
\begin{tabular}{p{14.5cm}}
\toprule
\textbf{Prompt for PopQA and RGBqa}: \\
Please answer the question based on the provided context and your own knowledge. Only include the answer in your response without any note or explanation, and try to be concise. Here is an example to help you know the format.
\#\#\#\# Example
Question: What holiday-themed Pop-Tart flavor does Pop-Tarts playfully suggest on their Instagram, eliciting mixed reactions?
The answer is: Gingerbread.
\#\#\#\# Example End
Paragraph: \{retrieved document\}
Question: \{question\} \\ \\
\textbf{Prompt for PubHealth}: \\ 
Please answer the statement is correct or not based on the provided paragraph and your own knowledge. Say true if it's correct; otherwise say false. Only include the answer in your response without any note or explanation, and try to be concise.  Paragraph: \{retrieved document\} Statement: \{question\} \\
\bottomrule
\end{tabular}
\caption{Prompts for different datasets. \{retrieved document\} and  \{question\} will be replaced with actual contents according to data.}
\label{tab:prompts}
\end{table*}

\section{Additional Details of Cognizant Experts}
\label{appendix:knowledge_setup}
In this subsection,  we provide a detailed description of using cognizant experts to predict whether the model's internal knowledge is sufficient. 


\paragraph{Dataset}
Given a RAG dataset $D$, we first divide it into $D_\text{pos}$  containing data points where the model answers correctly, and $D_\text{neg}$ consisting of data points where the model answers incorrectly. 
In our experiments, \textbf{Full-Set} is the combination of $D_\text{pos}$ and $D_\text{neg}$ and we use Full-Set as the evaluation data. 
To show the generalizability of cognizant experts, we also propose using 50-Shot, which consists of 50 samples randomly selected from the full set.

\paragraph{Method} Cognizant experts are identified through the Contrastive Expert Activation Inspection (CEAI).
First, based on Equation \ref{eq:p_d}, we use Full-Set or 50-Shot to calculate the contrastive activation probabilities $\Delta \mathbf{P} \in \mathbb{R}^{L \times N}$ for cognizant experts, where $L$ is the number of LLM's layers and $N$ is the number of experts per layer. 
With $\Delta \mathbf{P}$, we query LLM and get each sample's activation of an expert during LLMs' forward phase. The activation of experts allows us to compute a scenario score for each sample using Equation \ref{eq:score}.  
A positive scenario score indicates that the model’s internal knowledge is sufficient for the given question, indicating no external information is needed. Conversely, a negative score suggests knowledge is insufficient, necessitating external information. 
We only keep the Top-K and Bottom-K items in $\Delta \mathbf{P}$ to calculate the scenario score, Top-K and Bottom-K are hyperparameters that we search from 1 to 20 to find the optimal results. 

\paragraph{Evaluation} For the golden label of each sample, we consider questions that the model answers correctly as knowledge sufficiency and the questions that the model answers incorrectly as knowledge insufficiency. 
The metric is the F1 score between the prediction of scenario scores and the prediction of the model's real response.  
We use the F1 score because the data of knowledge sufficiency and knowledge insufficiency may be imbalanced, F1 score can provide a more fair comparison.
For 50-shot, we run experiments on five different seeds and reported the average metrics.

\paragraph{Baselines} We use Random Guess as a baseline for comparison, Random Guess randomly predicts the current scenario as either knowledge sufficiency or knowledge insufficiency with a 50\% probability. 
For Random Guess, we run experiments on five different seeds and reported the average metrics.

\section{Additional Details of Quality Experts}
\label{appendix:quality_setup}
In this subsection, we present a detailed methodology for employing quality experts to assess the quality of retrieved documents.

 Given a RAG dataset $D$, we construct the $D_\text{pos}$ by associating each question in $D$ with high-quality documents that contain the correct answers.  On the contrary, we provide low-quality documents for each question in $D$ to construct $D_\text{neg}$. 
 Low-quality documents are further divided into two categories: Distracting and Unrelated. 
 To get documents of different qualities, we can use exact matching and the similarity between questions and documents. 
 Specifically, we can retrieve documents from the external knowledge base, and select Top-k documents based on the similarity between these documents and questions. 
 Then, we can search for documents that contain answers, these documents are considered high-quality. 
 Documents that do not contain answers but have high similarity between questions are considered Distracting documents. 
 Documents that do not contain answers and have low similarity between questions are considered Unrelated documents. 
 RGBqa officially provides high-quality documents and Distracting documents. We use the retrieved for other questions as the Unrelated documents for this dataset.

 Once we build the $D_\text{pos}$ and $D_\text{neg}$, we can calculate the scenario score and predict the scenario for the given inputs. We also consider $D_\text{pos}$ and $D_\text{neg}$ as a full set and we randomly select 50 samples to construct 50-shot. 
 We also use Random Guess as a baseline for comparison, Random Guess randomly predicts the current scenario as either the high-quality retrieved document or the low-quality retrieved document with a 50\% probability.
 For Random Guess and 50-shot, we run experiments on five different seeds and reported the average metrics.
The detailed steps of calculating the score are similar to cognizant experts, which can be referred to in \S \ref{sec:activation_as_classifier} and Appendix \ref{appendix:knowledge_setup}.

\section{Additional Details of In-context Experts}
\label{appendix:retrieval_setup}

\paragraph{Dataset} 
Given a RAG dataset $D$, we construct $D_\text{pos}$ by providing the retrieved documents for each sample in $D$, and construct $D_\text{neg}$ by not providing any retrieved documents. 
Data in $D_\text{neg}$ use the same prompt as $D_\text{pos}$, which is shown in Appendix \ref{appendix:core_experts}. 
To mitigate bias introduced by varying text lengths, we pad samples in $D_\text{neg}$ to match the length of $D_\text{pos}$. For RGBqa, we use three high-quality retrieved documents. For PopQA, we use five retrieved documents with high similarity between questions.

\begin{table*}[t!]
    \resizebox{\linewidth}{!}{
    \begin{tabular}{lcccccccccccccccc}
    \toprule 
    \multirow{2}{*}{\textbf{Method}} & \multicolumn{3}{c}{\textbf{PubHealth}} & & \multicolumn{3}{c}{\textbf{PopQA}} & & \multicolumn{3}{c}{\textbf{$\text{RGBqa}_{mix}$}} & & \multicolumn{3}{c}{\textbf{BalanceQA}} \\
    \cline{2-4} \cline{6-8} \cline{10-12} \cline{14-16}
    & Acc & R-Score & R-Token & & Acc & R-Score & R-Token & & Acc &  R-Score & R-Token & & Acc & R-Score & R-Token  \\
    \midrule
    \textbf{No RAG} & 78.31 & 78.31 & 0 && 41.70 & 42.50 & 0 && 47.67 & 47.67 & 0 && 50.00 & 50.00 & 0\\
    \textbf{Always RAG} & 57.54 & 21.68 & 514274 && 53.60 & 57.50 & 519858 && 60.00 & 52.33 & 344916 && 50.00 & 50.00 & 214781\\
    \textbf{Random RAG} & 68.99 & 52.68 & 250540 && 48.60 & 49.40 & \textbf{261398} && 52.50 & 50.83 & \textbf{124746} && 49.00 & 47.75 & \textbf{113233}\\
    \midrule
    \textbf{Expert-RAG} \\
    w/ C & 80.95 & 80.14 & 76455 && 53.70 & \textbf{61.10} & 491900 && 62.17 & 68.33 & 221024 && 64.50 & \textbf{71.25} & 135474 \\
    w/ C\&Q & 80.95 & 80.14 & 76455 && \textbf{54.10} & 60.70 & 491900 && \textbf{63.67} & \textbf{70.66}& 199370 && \textbf{66.25} & 69.25 & 115819\\
    w/ C\&Q\&R & \textbf{81.36} & \textbf{80.14} & \textbf{76455} && 53.80 & 60.70 & 491900 && 62.83 & 70.66 & 199370 && 65.75 & 69.25 & 115819\\
    \bottomrule
    \end{tabular}
    }

\caption{Overall experiment results of Mixtral-8x22B on four datasets. Bold numbers indicate the best performance among RAG baselines. Acc is the Accuracy (task performance); R-Token is the token number of retrieved documents (inference cost). R-Score is the retrieval accuracy. For Expert-RAG, C means cognizant expert, Q means quality expert and R means in-context expert. w/ C\&Q\&R means using all three experts. }
\label{tab:main_results_22b}
\end{table*}

\paragraph{Method} As discussed in  \S \ref{sec:retrieval_expert}, we enhance the model's ability to utilize contextual information by manipulating expert activation within the model. Here we show the detailed steps of adjusting the model ability.  After obtaining the contrastive activation probabilities for in-context experts, we select the Top-k and the Bottom-k experts, which can be the core experts for using and not using contextual information. We search Top-k and the Bottom-k from [10,20,30,40,50] and report the best-desired performance.
To enhance the ability to utilize context, we enhance the selected Top-k (contextual) experts and inhibit the selected Bottom-k  (internal) experts. 
As mentioned in \S \ref{sec:moe}, the output of the MoE module is the sum of experts, which means increasing the weights of certain experts can improve their importance for output. As a result, for the Top-k expert, we force them to activate and increase their activation weights. For bottom-k experts, we force them not to be activated to reduce their influence on model output. 
Note that we do not increase the number of activated experts, each layer still activates the default number of experts. 
In our experiments, we set the weight of enhanced expert as 0.8 (the sum of the weights is 1) to empower its importance. 
If all activated experts are selected core experts, we set the weights of these experts equal. For example, if the model activates Top-2 experts and we want to enhance two experts in this layer, we set the weights of these experts equally as 0.5. 
For each input, we control the expert activation from generating the first token until the generation is complete.

\section{Additional Details of RAG Application}
\label{appendix:RAG_setup}
\paragraph{General Settings}
For PopQA, we utilize the same subset as described in \S \ref{sec:detecting_core_experts}, comprising 1,000 randomly selected samples.  For RGBqa, we create the $\text{RGBqa}_\text{mix}$, which consists of two types of data, one with three high-quality documents and another with three distracting low-quality documents, containing 600 samples in total. We use the full set of PubHealth, which contains 987 samples. For BalanceQA, we use PopQA,  $\text{RGBqa}_\text{mix}$, and PubHealth to query LLM and obtain various data types necessary for BalanceQA. We then randomly select 100 samples for each data type, forming a BalanceQA dataset with a total of 400 samples.
To identify the cognizant experts, we randomly select 50 samples from each dataset. 
For identifying quality experts, we follow the procedure outlined in \S \ref{sec:quality_expert} by selecting 50 random samples from RGBqa to construct the $D_\text{pos}$ and $D_\text{neg}$ datasets. The quality experts identified through this process are applied to all datasets. We use full-set to identify in-context experts. For each type of expert, we use the default method to implement and search hyperparameters, which are described in the respective section.
\begin{table}[h]
    \centering
    \resizebox{\linewidth}{!}{
    \begin{tabular}{lcccccc}
        \toprule
        \multirow{2}{*}{\textbf{Method}}  & \multicolumn{3}{c}{\textbf{TQA}} & \multicolumn{3}{c}{\textbf{NQ}} \\
        & EM $\uparrow$ & R-Score $\uparrow$ & R-Token $\downarrow$ & EM $\uparrow$ & R-Score $\uparrow$ & R-Token $\downarrow$ \\
        \midrule
        \textbf{No Retrieval} & 69.74 & \textbf{69.74} & 0 & 36.55 & 36.55 & 0 \\
        \textbf{Always Retrieval} & 70.74 & 30.25 & 984661 & 57.25 & 63.45 & 1042184 \\
        \textbf{Random Retrieval} & 70.37 & 50.44 & \textbf{493114} & 47.45 & 48.95 & \textbf{514777} \\
        \textbf{Expert-RAG} & \textbf{71.58} & 47.28 & 709726 & \textbf{57.50} & \textbf{64.10} & 1005172 \\
        \bottomrule
    \end{tabular}}
    
    \caption{Results on Natural Questions (NQ) and TriviaQA (TQA).}
    \label{tab:nq_tqa_result}
\end{table}

\paragraph{Results on additional dataset}
We conduct additional experiments on the Natural Questions (NQ) and TriviaQA (TQA) datasets. We randomly sample 200 queries from NQ and TQA to identify the experts, and evaluate our method on another 2000 randomly sampled queries. Each query is provided with 5 retrieved documents. The results are presented in the Table \ref{tab:nq_tqa_result}. The experimental results show that our method can also achieve good performance and efficiency on these authoritative datasets. This result indicates that our proposed method for identifying core experts can be applied to various types of data, and our identified three types of RAG-related core experts are widely present in different scenarios. Our proposed ARAG method based on core experts also has the ability to generalize to new scenarios. 

\begin{table*}[t!]
    \resizebox{\linewidth}{!}{
    \begin{tabular}{lcccccccccccccccc}
    \toprule
    \multirow{2}{*}{\textbf{Method}} & \multicolumn{3}{c}{\textbf{PubHealth}} & & \multicolumn{3}{c}{\textbf{PopQA}} & & \multicolumn{3}{c}{\textbf{$\text{RGBqa}_{mix}$}} & & \multicolumn{3}{c}{\textbf{BalanceQA}} \\
    \cline{2-4} \cline{6-8} \cline{10-12} \cline{14-16}
    & Acc $\uparrow$ & R-Score $\uparrow$ & R-Token $\downarrow$ & & Acc $\uparrow$ & R-Score $\uparrow$ & R-Token $\downarrow$ & & Acc $\uparrow$ & R-Score $\uparrow$ & R-Token $\downarrow$ & & Acc $\uparrow$ & R-Score $\uparrow$ & R-Token $\downarrow$  \\
    \midrule
    \textbf{EAR} & 54.40 & 49.94 & 194727 & & 49.50 & 67.90 & 499653 & & 58.50 & 63.33 & 344916 & & 53.50 & 56.00 & 106511 \\
    \textbf{Self-RAG (No-Retrieval)} & 70.01 & \textbf{70.01} & 0 & & 23.50 & 23.50 & 0 & & 24.33 & 24.33 & 0 & & 50.00 & 50.00 & 0 \\
    \textbf{Self-RAG (threshold=0.3)} & \textbf{71.93} & 29.98 & 514276 & & 45.30 & \textbf{76.40} & 519331 & & 59.83 & \textbf{75.00} & 342491 & & 50.05 & 50.00 & 206165 \\
    \textbf{Self-RAG (threshold=0.4)} & 72.54 & 35.96 & 466750 & & 40.80 & 64.10 & 412535 & & 52.33 & 64.33 & 252915 & & 52.75 & 50.75 & 171477 \\
    \textbf{Self-RAG (threshold=0.5)} & 70.61 & 68.08 & \textbf{47289} & & 25.70 & 28.50 & \textbf{75728} & & 26.83 & 28.00 & \textbf{17879} & & 50.25 & 49.50 & 19146 \\
    \textbf{Expert-RAG (7B)} & 58.05 & 63.02 & 369555 & & \textbf{49.90} & 72.20 & 439033 & & \textbf{60.67} & 66.88 & 297062 & & \textbf{59.00} & \textbf{64.70} & \textbf{104507} \\
    \bottomrule
    \end{tabular}
    }
\caption{Comparison with other adaptive RAG methods. Bold numbers indicate the best performance. Acc is the Accuracy (task performance); R-Token is the token number of retrieved documents (inference cost). R-Score is the retrieval accuracy.}
\vspace{-0.4cm}
\label{tab:other_baseline}
\end{table*}
\paragraph{Results of Mixtral-8x22B} 
We also apply Expert-RAG to the Mixtral-8x22B and show the results in Table \ref{tab:main_results_22b}. 
We observed that our method demonstrates greater advantages on the 8x22B model compared to the 8x7B model, particularly on the PubHealth and BalanceQA datasets. 
On PubHealth, the Acc score of Always RAG is 10 points lower than that of No RAG. This indicates that when the internal knowledge of the model is sufficiently strong, retrieved documents may mislead LLM and result in a decline in performance. A similar phenomenon is also noted in \cite{asai2024selfrag}. 
However, Expert-RAG mitigates the inference costs and negative impacts associated with retrieval on such datasets, achieving optimal performance. Additionally, on BalanceQA, our method yields a more substantial performance improvement compared to the 8x7B model, suggesting that more diverse contrastive activation patterns can achieve better results. Additionally, we can find that in-context experts are not always helpful. This is because quality experts cannot filter all low-quality documents, there is still some low-quality documents that can mislead LLM. Enhancing the ability of using information from the context may have negative effect in such scenarios. 
In summary, the results of Mixtral-8x7B-instruct0.1 and Mixtral-8x22B-instruct0.1 together demonstrate the effectiveness of our method and the assistance of expert activation for RAG.

\paragraph{Comparison with other adaptive RAG baselines} 
We compare our method with two adaptive RAG methods: Entity-based Adaptive-Retrieval (EAR) \cite{mallen-etal-2023-trust} and Self-RAG \cite{asai2024selfrag}. EAR uses the popularity of entity words in the question to determine whether retrieval is necessary, while Self-RAG fine-tuning LLM to enable the model to perform adaptive RAG. The experimental results are shown in the Table \ref{tab:other_baseline}. 
Compared to entity-based adaptive retrieval, our method achieves better performance and efficiency on all datasets. This is because our method takes into account the model's true internal state (expert activation) and can more accurately reflect the internal knowledge of the model, whereas EAR only uses entity popularity to estimate knowledge sufficiency. Compared to Self-RAG, our method achieved better Acc and R-Token on PopQA, $\text{RGBqa}_{Mix}$, and BalanceQA. However, we want to point out that Expert-RAG and Self-RAG use different base model. Self-RAG requires fine-tuning on 150k data to acquire adaptive RAG capabilities, we are not able to align with it during the rebuttal period due to the time and resources constrain. The different knowledge and capabilities of different models may be the one reason for the divergent results across different datasets. On PubHealth, our method performed worse than Self-RAG. We speculate this is also due to SFT making Self-RAG more suitable for this particular task. This is evidenced by the fact that EAR and Expert-RAG, without SFT, can only achieve an Acc score of around 50, while Self-RAG (No Retrieval) reaches an Acc of 70.01 on PubHealth even without using retrieval documents. And Self-RAG (threshold=0.4) with adaptive retrieval enabled achieve an Acc of 72.54. This suggests that Self-RAG's advantage on PubHealth primarily stems from the model's inherent capabilities rather than from adaptive retrieval. In summary, our method has shown advantages in most data, demonstrating its effectiveness in the ARAG scenario.

\end{document}